\documentclass[letterpaper, 10 pt, conference]{ieeeconf}  

\IEEEoverridecommandlockouts
\usepackage{cite}
\usepackage{subfigure}
\usepackage{amsmath,amssymb,amsfonts}
\usepackage{algorithmic}
\usepackage{graphicx}
\usepackage{textcomp}
\usepackage{xcolor}
\usepackage{balance}
\def\BibTeX{{\rm B\kern-.05em{\sc i\kern-.025em b}\kern-.08em
    T\kern-.1667em\lower.7ex\hbox{E}\kern-.125emX}}
\usepackage{array}

\usepackage{float}
\usepackage{stfloats}

\usepackage{listings}
\usepackage{xcolor}
\lstdefinelanguage{json}{
    basicstyle=\ttfamily\small,
    numbers=left,
    numberstyle=\tiny,
    stepnumber=1,
    numbersep=5pt,
    showstringspaces=false,
    breaklines=true,
    frame=single,
    backgroundcolor=\color{gray!5},
    string=[s]{"}{"},
    morestring=[b]',
    literate=
     *{0}{{{\color{blue}0}}}{1}
      {1}{{{\color{blue}1}}}{1}
      {2}{{{\color{blue}2}}}{1}
      {3}{{{\color{blue}3}}}{1}
      {4}{{{\color{blue}4}}}{1}
      {5}{{{\color{blue}5}}}{1}
      {6}{{{\color{blue}6}}}{1}
      {7}{{{\color{blue}7}}}{1}
      {8}{{{\color{blue}8}}}{1}
      {9}{{{\color{blue}9}}}{1}
}
\lstdefinelanguage{json}{
    basicstyle=\ttfamily\small,
    numbers=left,
    numberstyle=\tiny,
    stepnumber=1,
    numbersep=5pt,
    showstringspaces=false,
    breaklines=true,
    frame=lines,
    backgroundcolor=\color{gray!5},
    literate=
     *{:}{{{\color{blue}:}}}{1}
      {,}{{{\color{red},}}}{1}
      {\{}{{{\color{orange}{\{}}}}{1}
      {\}}{{{\color{orange}{\}}}}}{1}
      {[}{{{\color{orange}{[}}}}{1}
      {]}{{{\color{orange}{]}}}}{1},
}

\begin{document}

\bstctlcite{IEEEexample:BSTcontrol}

\title{\LARGE \bf Toward Accurate Long-Horizon Robotic Manipulation: \\ Language-to-Action with Foundation Models via Scene Graphs
}

\author{Sushil Samuel Dinesh and Shinkyu Park 
\thanks{The work was supported by funding from King Abdullah University of Science and Technology (KAUST).}
\thanks{The authors are with the Department of Electrical and Computer Engineering, King Abdullah University of Science and Technology (KAUST), Thuwal, 23955, Saudi Arabia. {\tt \{sushilsamuel.dinesh, shinkyu.park\}@kaust.edu.sa}}
}

\IEEEoverridecommandlockouts
\overrideIEEEmargins

\maketitle

\begin{abstract}

This paper presents a framework that leverages pre-trained foundation models for robotic manipulation without domain-specific training. The framework integrates off-the-shelf models, combining multimodal perception from foundation models with a general-purpose reasoning model capable of robust task sequencing. Scene graphs, dynamically maintained within the framework, provide spatial awareness and enable consistent reasoning about the environment. The framework is evaluated through a series of tabletop robotic manipulation experiments, and the results highlight its potential for building robotic manipulation systems directly on top of off-the-shelf foundation models.
\end{abstract}


\section{Introduction}
The primary motivation of this work is to develop a framework that integrates multiple \textit{foundation models}---pre-trained large models---to enable perception, planning, and execution without requiring dedicated end-to-end training or fine-tuning, while maintaining high accuracy. In this framework, users provide verbal commands specifying the desired objective, and the robot perceives the environment, generates a plan, and executes planned tasks directly from this high-level input.

\begin{figure}[t] 
  \centering
  \includegraphics[width=1\columnwidth]{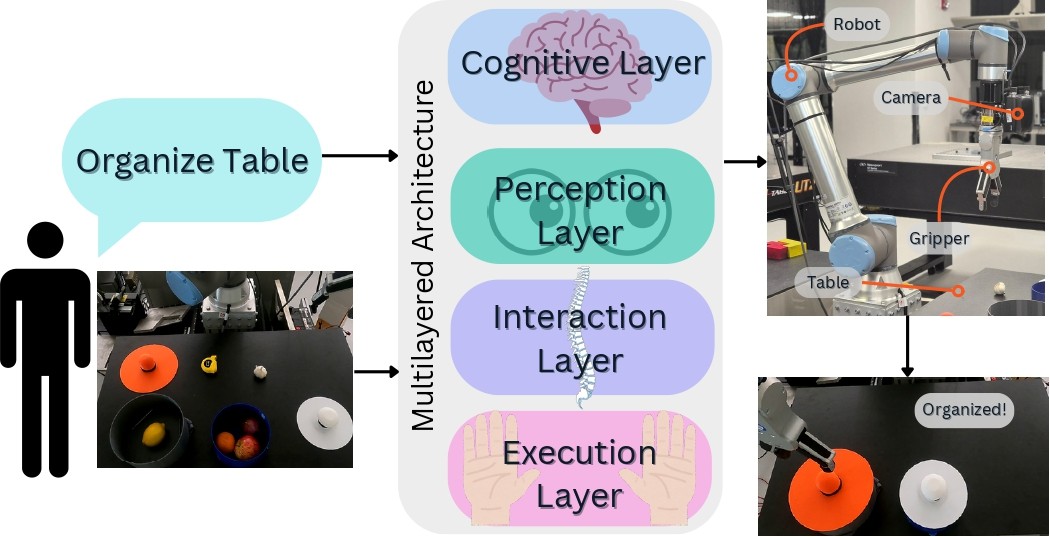}
  \caption{Overview of the Proposed Framework: The framework is organized into multiple layers, each with distinct capabilities, and is designed to translate high-level natural language commands from the user into an executable sequence of robot actions.}
  \label{fig:framework_overview}
  \vspace{-1.5em}
\end{figure}

Our framework, as illustrated in Fig.~\ref{fig:framework_overview}, is built on a layered architecture, where each layer is governed by a specialized model and the overall system emerges through their structured interconnection. Many existing approaches, in contrast, employ these models in more constrained ways---for example, as \textit{Large Language Model (LLM)}-based planners without basing their reasoning in spatial understanding \cite{robogpt, instruct2act, gpt_4v_microsoft}, as \textit{Vision-Language Models (VLMs)} that generate trajectories from image overlays but are often error-prone \cite{pivot, moka, rekep}, or as \textit{Vision-Language-Action (VLA) models} that directly map images and language to robot actions but require massive amounts of training data \cite{rt1, rt2, openx}.

In the proposed framework, multiple foundation models are interconnected to address different stages of workflow. In particular, the LLM interprets the user's request, the VLM provides perception, and a reasoning model generates a detailed task sequence. Each task in the sequence is then executed by a model that integrates a motion planner with a motor controller. Beyond the layered architecture, we incorporate \textit{scene graphs} \cite{scenegraphs_survey} to provide structured representations of the environment, allowing the models to inform their reasoning more effectively and interpret the workspace with greater accuracy. We evaluate the framework through a series of experiments, ranging from simple object relocation to puzzle solving and long-horizon tasks, demonstrating its versatility and effectiveness.

\subsection{Comparative Review of Related Work}

\textit{LLMs for Planning and Sequencing: }
Early efforts to integrate LLMs into robotics generated robot actions by parsing outputs into structured formats like ``\texttt{\small [Action Name] [Object Name] to [Position Name]}'', but were limited to simple tasks in simulation \cite{robogpt}. Subsequent extensions introduced prompt engineering and custom functions, but these approaches still relied on explicitly provided object details and lacked autonomous scene understanding \cite{gpt_empowered_long_step_control}.

Subsequent efforts advanced toward multi-model frameworks, for example YOLO-based object recognition with waypoint extraction from human demonstrations \cite{gpt_4v_microsoft}. The study of \cite{xu2023} employed exemplars and rule-based systems, improving instruction following but remaining tied to code-based execution. Similarly, \cite{instruct2act} demonstrated that LLMs could not introduce new logic (e.g., failure handling) beyond provided examples, motivating direct LLM-driven tool handling (function calling) in a feedback loop.

Collectively, these studies underscore a key limitation: while LLMs excel at symbolic reasoning, they remain weak in integrating reasoning within the physical world.

\textit{VLMs for Perception and Spatial Reasoning: }
Several studies have explored the use of overlays---such as arrows or grid-based keypoints---to guide trajectory generation using VLMs \cite{pivot, moka}. The study of \cite{rekep} introduced keypoint-based spatial reasoning by deriving 3D coordinates to form constraint functions and sub-goals. While this provided effective for tasks such as cloth folding, it struggled to generalize to long-horizon planning, and overall experimental success remained limited.

Other works emphasized dynamic scene understanding. For example, \cite{nlmap} used  LLMs to generate object proposals from natural language but remained tied to feature-based representations without explicit spatial modeling. A more complex pipeline explored in \cite{gao_thesis_9} incorporated Visual Question Answering (VQA), but still depended on task-specific datasets, bounding box detection, and a fine-tuned InstructBLIP model \cite{instructblip}.

In summary, perception in these approaches often remained decoupled from high-level reasoning, leading to brittle pipelines vulnerable to error propagation.

\textit{Scene Graphs and Affordance-based Reasoning: }
SayPlan \cite{sayplan} explored scene-graph-based manipulation and planning with affordances for mobile robots, emphasizing semantic search in large multi-room environments. Hydra \cite{hydra} extended this direction by developing real-time, hierarchical scene graph construction through semantic segmentation, though without considering affordances.

SayCan \cite{saycan} grounded LLMs in robotics using affordance functions, enabling language-guided task planning with real-world feasibility; however, their approach depended on a large dataset of $68,000$ teleoperated actions across $10$ robots. VoxPoser \cite{voxposer} derived affordance maps from VLMs to produce 3D representations for trajectory optimization and obstacle avoidance, but it lacked long-horizon planning capabilities.

Overall, while scene graphs and affordances provide strong mechanisms for semantic contextualization, they have not yet been unified with the reasoning depth of foundation models to support robust task sequencing in robots.

\textit{Action Generation and Learning-Based Approaches: }
Recent VLA models have significantly advanced robot action generation. PaLM-E \cite{palme} introduced short-horizon embodied models, while RT-1 \cite{rt1} and RT-2 \cite{rt2} scaled training with large datasets, enabling closed-loop control and language-conditioned reasoning but with limited generalization. GR00T \cite{gr00t} combined a reasoning-capable VLM with a fast diffusion-based motion generator, and Open-X \cite{openx} extended this direction through cross-robot training across $22$ robots and $160,000$ tasks. 

Despite these advances, VLA models remain highly data-hungry and struggle to generalize to long-horizon tasks.

\textbf{Research Positioning: }
Our work addresses these gaps by proposing a structured, layered framework that balances the reasoning strengths of LLMs with the perceptual capabilities of VLMs, organized through scene graphs and executed via conventional motion planning and control. Unlike data-intensive VLA approaches, our framework minimizes the need for task-specific datasets or fine-tuning, thereby reducing the engineering burden of developing dedicated long-horizon robot action models. At the same time, the framework integrates precise object grounding in perception through a state-of-the-art VLM, primary task planning via a powerful LLM, contextual scene understanding through LLM-VLM dialogue, and persistent spatial reasoning supported by online scene graph updates and task execution by a faster non-reasoning LLM in a feedback loop.

Through experiments demonstrating the successful execution of increasingly complex tasks, this work shows foundation models, when integrated with structured world representations, can effectively bridge the gap between language and action in robotics. The framework positions itself as a middle ground: more generalizable and adaptable than dataset-heavy VLA models, yet more sophisticated and spatially grounded than purely symbolic LLM planners. This dual advantage highlights a promising path toward scalable, adaptable, and semantically informed robotic manipulation.

\subsection{Paper Organization}

Section~\ref{sec:framework_design} presents the framework design, detailing the foundation models assigned to each layer of the architecture; the excerpts of the system prompt commands---that is, the preliminary instructions the LLM is expected to follow---are provided in the Appendix. Section~\ref{sec:experiment} validates the effectiveness of the framework across a range of manipulation tasks. Finally, Section~\ref{sec:conclusion} concludes the paper with a summary and directions for future research.

\section{Framework Design and Implementation} \label{sec:framework_design}

\subsection{Hardware Setup for Target Robotic Systems}
Our framework is designed for robotic manipulators equipped with a vision system capable of capturing image data from the workspace. In our experiment setup, as depicted in Fig.~\ref{fig:framework_overview}, we employ a UR10e collaborative robotic arm ($6~DOF$, $1300~mm$ reach, $12.5~kg$ payload) with joint velocity limits for safety, paired with an OnRobot RG6 gripper for handling diverse objects. Perception is supported by a Zivid 2 wrist-mounted 3D camera, which provides high-resolution RGB-D data for reliable scene understanding.

\subsection{Scene Graph}

The scene graph, as depicted in Fig.~\ref{fig:scenegraph}, serves as the central knowledge base, encoding spatial relationships, object properties, and semantic details in a format accessible to both foundation models and the motion planner. Implemented with the NetworkX library, it employs a hierarchical JSON representation that balances readability and efficiency. Its design draws inspiration from \cite{sayplan}. An example of scene graph generation and update is provided in the Supplementary Materials.

Each node represents an entity, ranging from the root node denoting the \textit{entire workspace} to individual objects such as an apple or a ball. Nodes capture affordances (e.g., \textit{pickable}), positions, coordinates, and domain knowledge, while edges encode containment relations. Together, they enable the LLM to reason about both physical and semantic constraints during task planning.

Scene graphs can be generated automatically (via GPT-4.1 and Qwen-2.5VL) or manually for complex setups requiring precise ground truth. In our framework, they are dynamically updated by the LLM to ensure consistency as tasks progress: perception outputs refine object positions, while manipulation actions (e.g., placing an apple in a box) update coordinates and relationships accordingly. Incremental updates are supported through targeted edits, allowing waypoints to be added, properties refined, or user-provided information integrated seamlessly. 

\subsection{Framework Capabilities}
Our framework enables robots to interpret natural-language instructions and generate long-horizon manipulation plans without task-specific training. Leveraging a 3D camera and VLM, it can perceive and localize novel objects in 3D space, compute precise coordinates in the robot’s frame for grasping and placement, and conduct VQA-style queries between the LLM and VLM to answer semantic questions about the scene (e.g., ``\texttt{\small do you see some blue object?}''). A persistent scene graph maintains object properties and relations, supporting reasoning to infer implicit goals. A reasoning-oriented model produces multi-step, constraint-aware task sequences, while a faster, execution-oriented LLM coordinates motion primitives and continuously updates the scene graph during task execution.

\subsection{Layered Architecture Design}

\begin{figure*}[t] 
  \centering
  \subfigure[] {
  \includegraphics[angle=0, height=.388\textwidth]{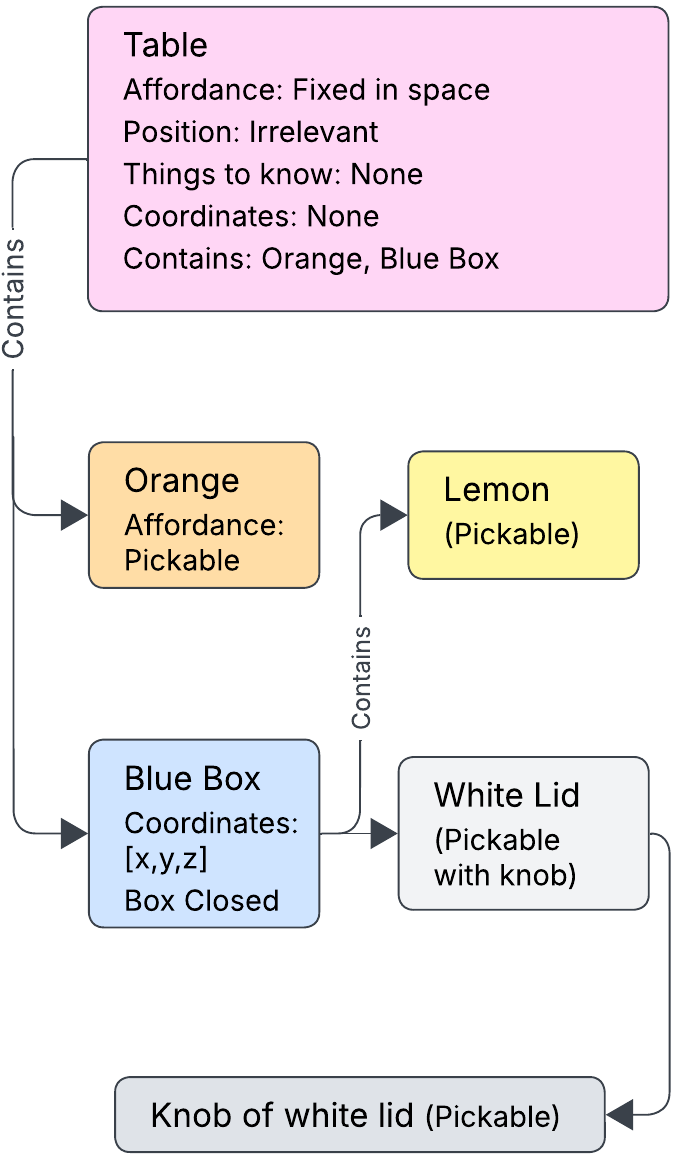}
  \label{fig:scenegraph}
  }
  \quad
  \subfigure[] {
  \includegraphics[width=.71\textwidth]{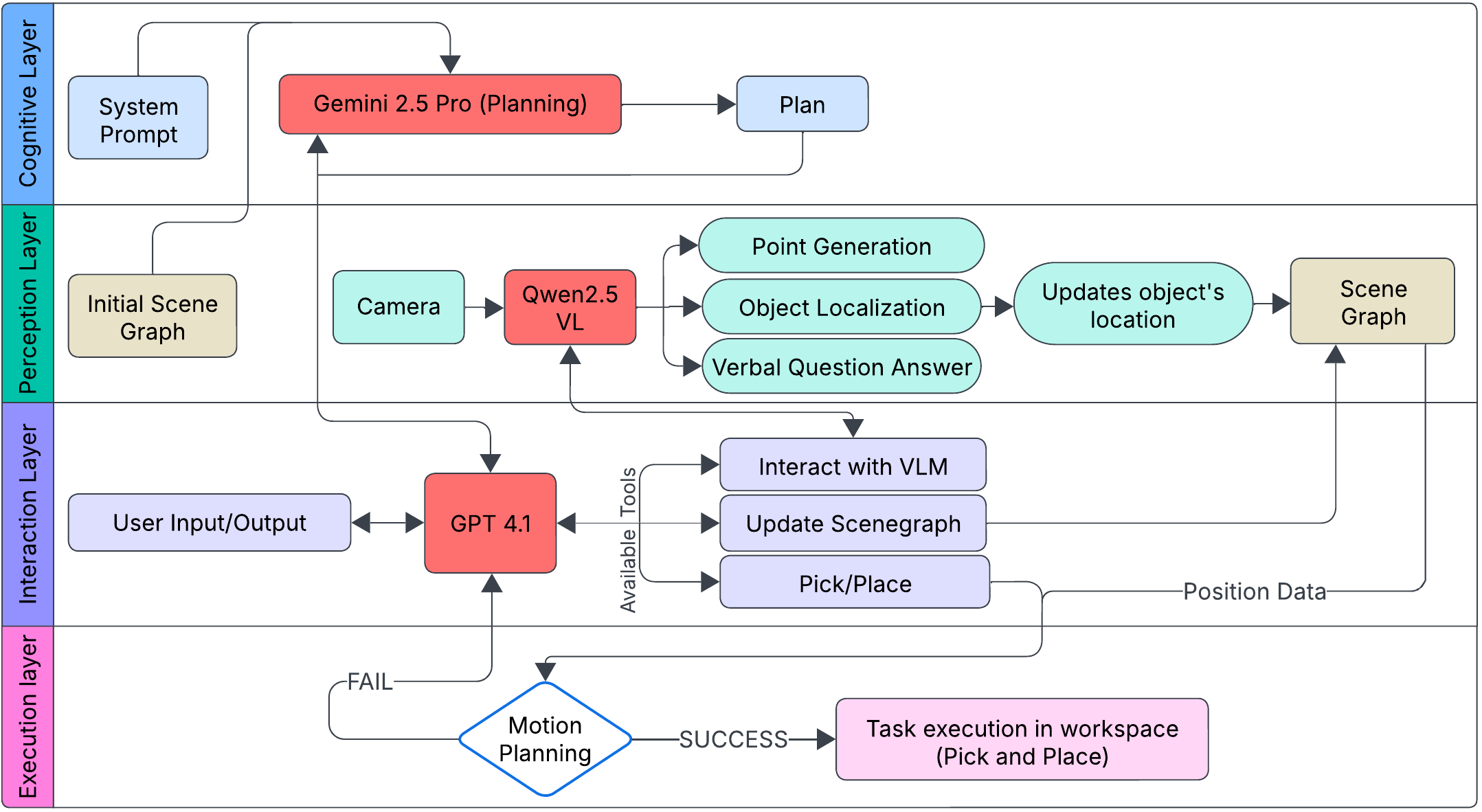}
  \label{fig:system_architecture}
  }
  
  \caption{(a) Scene Graph Structure. (b) System Architecture. The layers are organized in a bottom-up hierarchy. \textit{Execution Layer:} Relies on a conventional motion planner and controller to ensure robust and precise object manipulation. 
  \textit{Interaction Layer:} Utilizes a powerful, non-reasoning model to interpret user instructions and coordinate task execution.
  \textit{Perception Layer:} Incorporates a VLM with RGB-D input from a 3D camera to provide spatial understanding, object localization, and semantic scene descriptions
  \textit{Cognitive Layer:} Employs a reasoning model for advanced long-horizon planning and decision-making. }
  \vspace{-1.5em}
\end{figure*}

The proposed framework integrates foundation models with robotic hardware to enable natural language-driven task planning and execution without domain-specific training. Figure~\ref{fig:system_architecture} illustrates the layered architecture of the framework, where each layer is implemented as a modular package within the Robot Operating System 2 (ROS2) framework \cite{ros2}. The Appendix summarizes the prompts used for GPT-4.1, Qwen2.5-VL, and Gemini 2.5 Pro, the communication between these models, the tools available to GPT-4.1, and the failure-handling procedure used.

In the following, we present detailed explanations of each layer's functionality and design considerations.

\subsubsection{Execution Layer}
The execution layer translates planned tasks into safe and precise robot motions. Nvidia cuRobo’s GPU-accelerated motion planner generates collision-free trajectories that account for robot kinematics, grasp constraints, and workspace boundaries. Coordinated control of the arm and gripper ensures stable pick-and-place operations, while ROS2 integration provides seamless communication with higher-level layers \cite{curobo}.

\subsubsection{Interaction Layer}
The interaction layer requests, receives, and executes the high-level task plan produced by the cognitive layer while coordinating with the other layers. Upon receiving the task sequence, it orchestrates the available tools---function calls such as object-manipulation primitives (pick-and-place), perception queries, and scene-graph updates---step by step. These tools are provided to the LLM as callable functions, with each tool call corresponding to the execution of one such function.

At its core, OpenAI’s GPT-4.1—an evolution of GPT-4 \cite{gpt4}—provides natural-language interpretation and function calling, forming a seamless bridge between the user and the robot’s subsystems. During execution, the interaction layer monitors the return outputs of each function call, dynamically re-plans or adjusts the sequence when necessary.

In addition to internal coordination, it communicates directly with the user to provide status updates, request clarifications, or deliver results. This dual role---mediating between layers and interacting with the user---ensures that user intent is reliably realized while maintaining an intuitive, dialogue-driven interface.

\subsubsection{Perception Layer}
The perception layer connects visual input with semantic and geometric understanding, providing the spatial context required for robotic manipulation. At its core, it integrates the LLM–VLM dialogue system, where the LLM issues structured queries to the VLM. This interaction enables the robot to ``see'' and reason about its environment in a language-driven workflow, enabling perceptual capabilities that traditional vision pipelines would not achieve as fluidly.

Perception tasks follow along two main pathways:
\begin{itemize}

\item \textbf{Bounding Box Localization:} For accurate object localization, GPT-4.1 queries Qwen2.5-VL to generate bounding boxes around target objects. The process begins with a binary classification query (``\texttt{\small Is the target object present?}'') to avoid hallucinated detections. For confirmed objects, Qwen2.5-VL outputs pixel-space bounding boxes, which are converted into 3D coordinates using depth data from the robot's 3D camera and calibrated transformations. Multi-view reconstruction then merges point clouds collected from different camera viewpoints, removes the table plane and outliers, and applies clustering to segment objects. Each cluster is matched to the VLM-derived bounding box point cloud, and the centroid of the matched cluster is returned as the object's precise 3D location.

\item \textbf{Specific Point Queries:} For tasks such as finding a free placement spot, the framework employs targeted point queries. Qwen2.5-VL directly outputs pixel coordinates $(x, y)$ on the 2D image plane, which are projected into 3D space via depth map. This bypasses multi-view reconstruction while still ensuring precise placement coordinates.
\end{itemize}

Qwen2.5-VL is chosen for its strong spatial reasoning and object-grounding capabilities. Beyond object recognition, it aligns visual features with spatial queries to produce accurate localizations, even in cluttered or ambiguous scenes. Unlike conventional detectors, it requires no retraining on domain-specific datasets, making it highly adaptable. For instance, it can generate bounding boxes or placement points for novel objects and arrangements outside its training distribution, demonstrating robust generalization. This flexibility is particularly valuable in robotic manipulation, where robots often encounter unfamiliar objects and spatial configurations \cite{qwen2.5vl}.

By combining robust bounding-box detection, flexible point querying, and natural language-driven perception, the perception layer equips the robot with a general-purpose, adaptive understanding of its workspace. This enables reliable spatial awareness for complex manipulation tasks while maintaining real-time performance.

\subsubsection{Cognitive Layer}
The cognitive layer is dedicated to reasoning and planning. Google's Gemini 2.5 Pro Preview \cite{gemini2.5} generates multi-step task strategies by integrating information from the scene graph, user goals, and tool definitions i.e., the precise descriptions of each tool’s purpose, input parameters, and return values. Since reasoning models are relatively slow, Gemini 2.5 Pro Preview is not suited for the interaction layer. By separating planning (Gemini 2.5 Pro Preview) from execution via the interaction layer (GPT-4.1), the framework combines advanced reasoning with fast, reliable control.

Gemini 2.5 Pro Preview was selected for its strong performance across key dimensions on the LiveBench LLM benchmark \cite{livebench} at the time of this study. These include reasoning (essential for spatial and multi-step task planning), language comprehension (crucial for interpreting nuanced natural language instructions), instruction following (necessary for adhering to system prompt rules), and data analysis (particularly relevant for scene graph interpretation).

\section{Experimental Evaluation} \label{sec:experiment}

Experiments are designed to evaluate whether our foundation-model-driven framework can generalize across diverse tasks without task-specific training. As noted in \cite{general_purpose_robots_survey}, there are currently no established benchmarks for assessing the performance of such frameworks. Therefore, rather than pursuing narrow benchmark scores, we demonstrate the proposed framework's capabilities in perception, reasoning, and planning across a range of settings, including those involving ambiguous or underspecified user requests.

Three classes of experiments are designed to systematically evaluate the framework's performance and generalizability in complex real-world scenarios, using four key metrics as evaluation criteria:
   
\begin{itemize}

\item \textit{Planning Feasibility ($PF$):} The feasibility and goal-alignment of the generated task sequences from the cognitive layer.

\item \textit{Task Completion Rate ($TCR$):} The percentage of tasks successfully executed.

\item \textit{Scene Graph Handling ($SGH$):} The correctness of updates made to the world model by the end of each experiment.

\end{itemize}

To simplify system implementation and the experiment design, we make the following assumptions:

\begin{itemize}

\item 
Since dexterous object manipulation is not the focus of this work, we restrict the scenario to the robot manipulating objects through pick-and-place primitives, with grasping constrained to the centroid of each object.

\item API errors and network interruptions are excluded from the failure cases.

\item While the motion planner accounts for self-collision and static obstacles (e.g., the table), it does not model collisions with manipulatable objects.

\item Our focus is on how the framework accurately updates the scene graph rather than generating it from scratch. Accordingly, initial scene graphs are either provided manually or produced by the LLM at the beginning of each experiment. In particular, the LLM generates the initial graphs for Exps.~I-A, I-B, and II-A. For the remaining experiments, the graphs are manually created from scratch (Exp.~II-B) or manually created and subsequently modified through simple object or attribute additions and relabeling (Exps.~III-A$\sim$III-C).

\end{itemize}

\subsection{Experiment I: Testing Fundamental Capabilities}

\begin{figure*}[t]
  \centering
  \subfigure[] {
  \includegraphics[trim={15 0 10 0}, clip, width=0.145\textwidth]{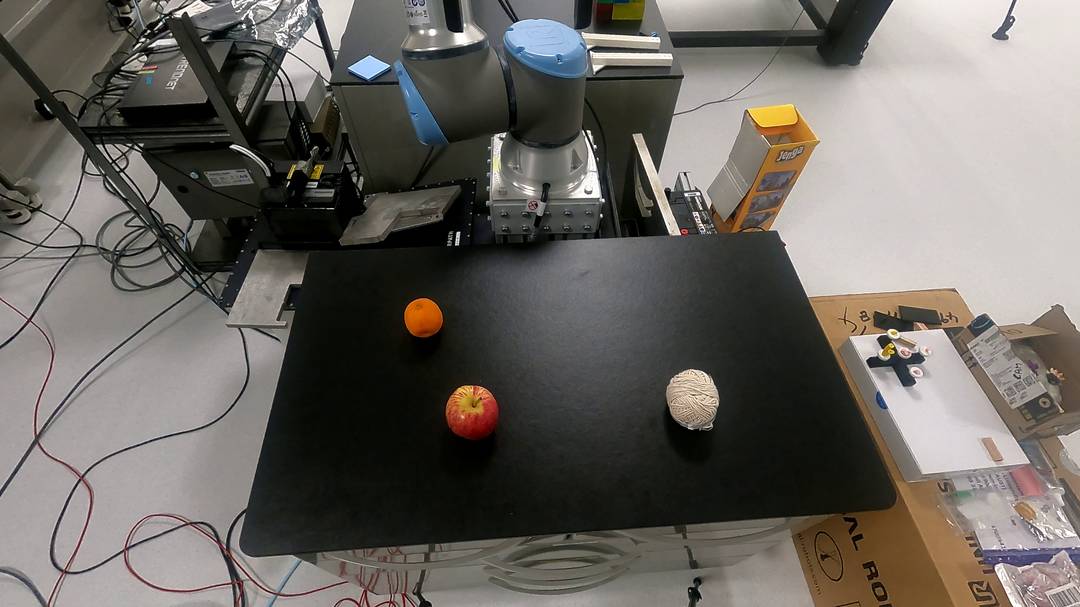}
  \label{fig:experiment_1_1}
  }
  \subfigure[] {
  \includegraphics[trim={15 0 10 0}, clip, width=0.145\textwidth]{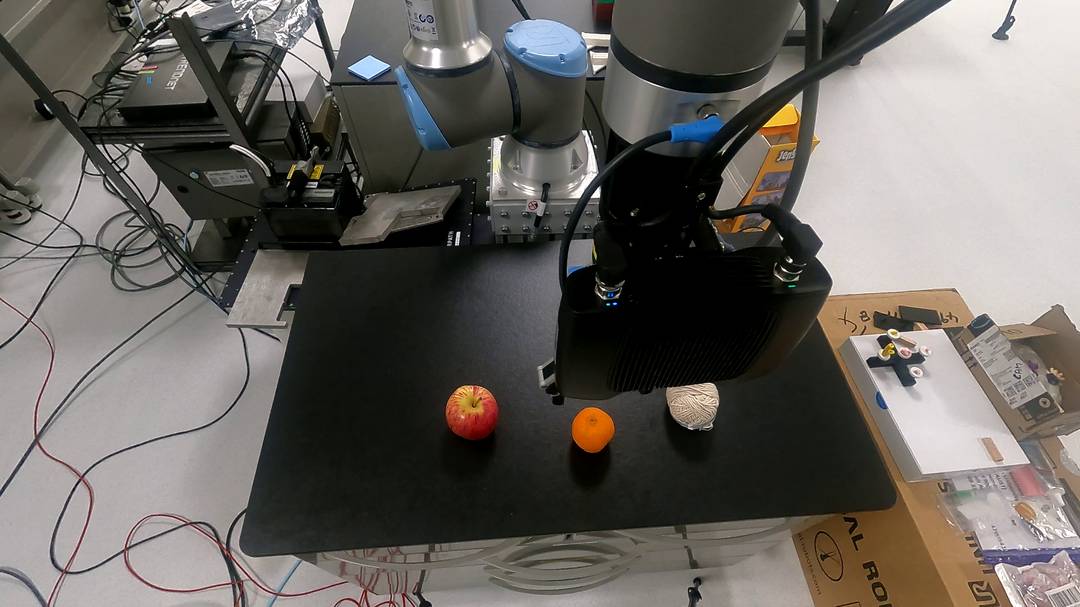}
  \label{fig:experiment_1_2}
  }
  \subfigure[] {
  \includegraphics[trim={5 0 5 0}, clip, width=0.145\textwidth]{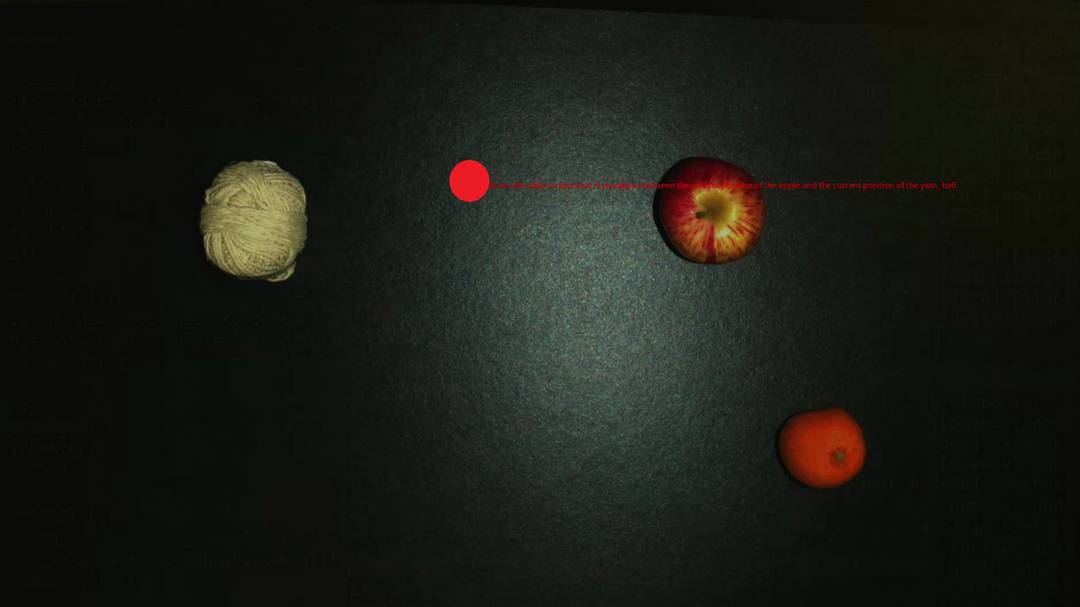}
  \label{fig:experiment_1_3}
  }
  \subfigure[] {
  \includegraphics[trim={15 0 10 0}, clip, width=0.145\textwidth]{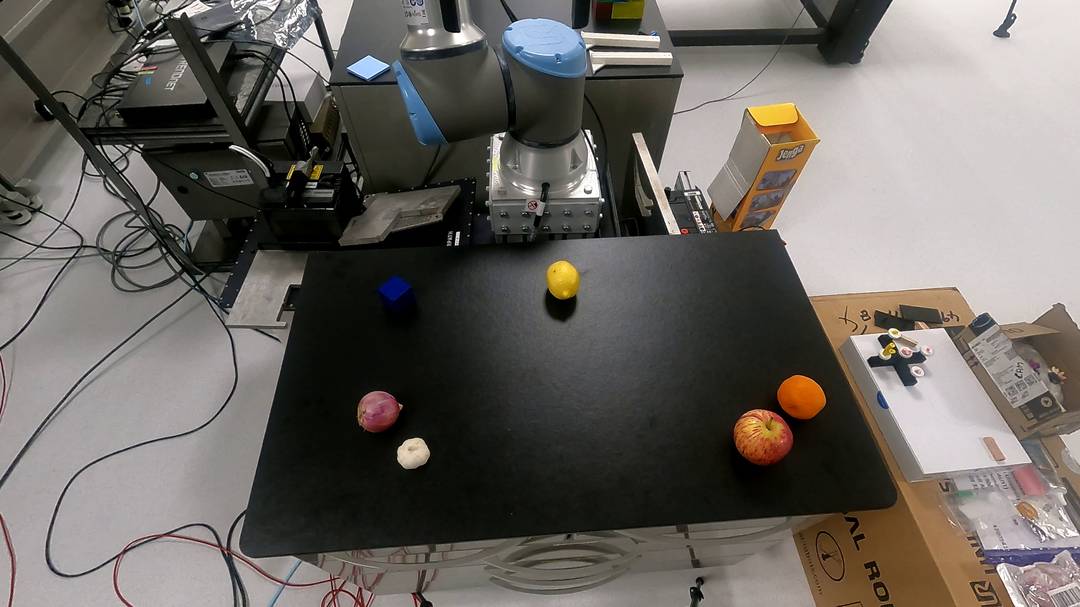}
  \label{fig:experiment_1_4}
  }
  \subfigure[] {
  \includegraphics[trim={15 0 10 0}, clip, width=0.145\textwidth]{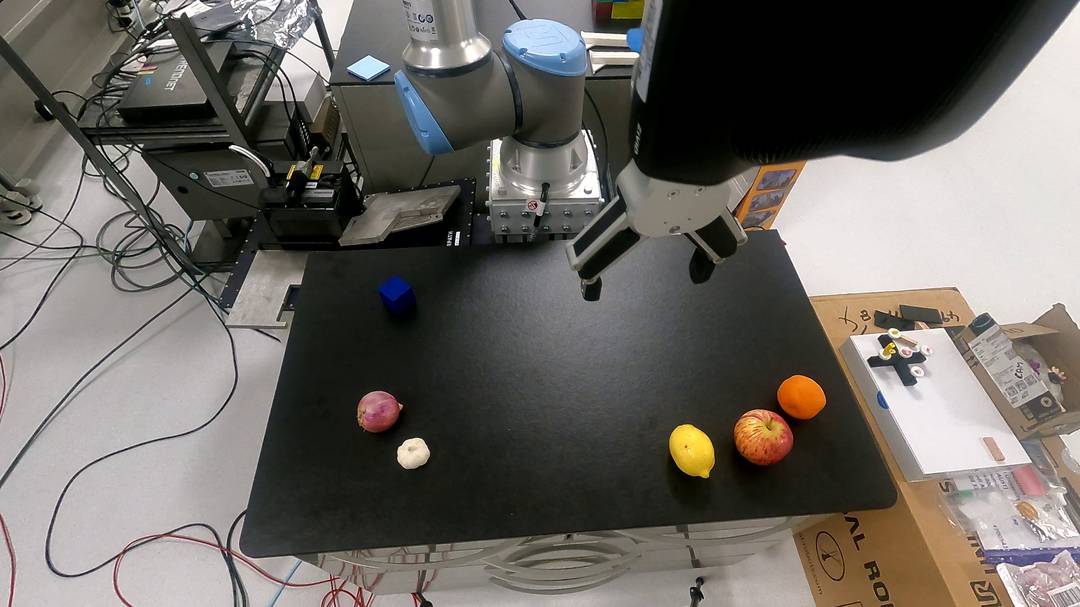}
  \label{fig:experiment_1_5}
  }
  \subfigure[] {
  \includegraphics[trim={0 0 10 0}, clip, width=0.145\textwidth]{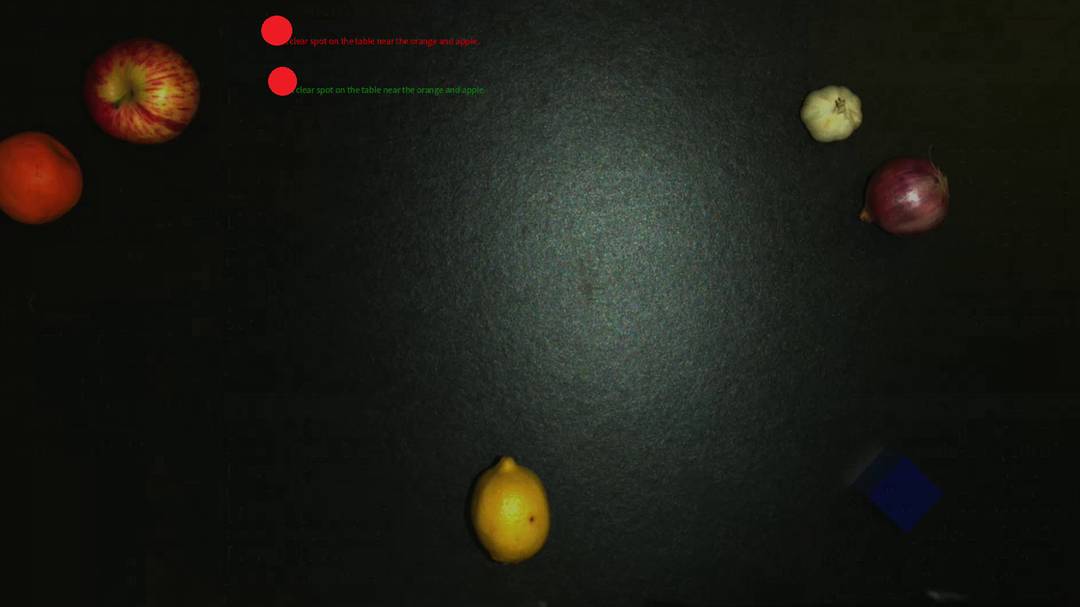}
  \label{fig:experiment_1_6}
  }
  \vspace{-.5em}
  \caption{Experiments I-A and I-B. I-A: (a)–(b) The orange moves from its initial position to between the apple and yarn; (c) shows the VLM-identified point satisfying the ``in-between'' condition. I-B: (d)-(e) The lemon shifts toward its correct cluster; (f) shows feasible points obtained from the VLM.}
  \vspace{-.5em}
  \label{fig:experiment_I-A_I-B}
\end{figure*}

\begin{figure*}[t]
  \centering
  \subfigure[] {
  \includegraphics[trim={0 0 0 0}, clip, width=0.225\textwidth]{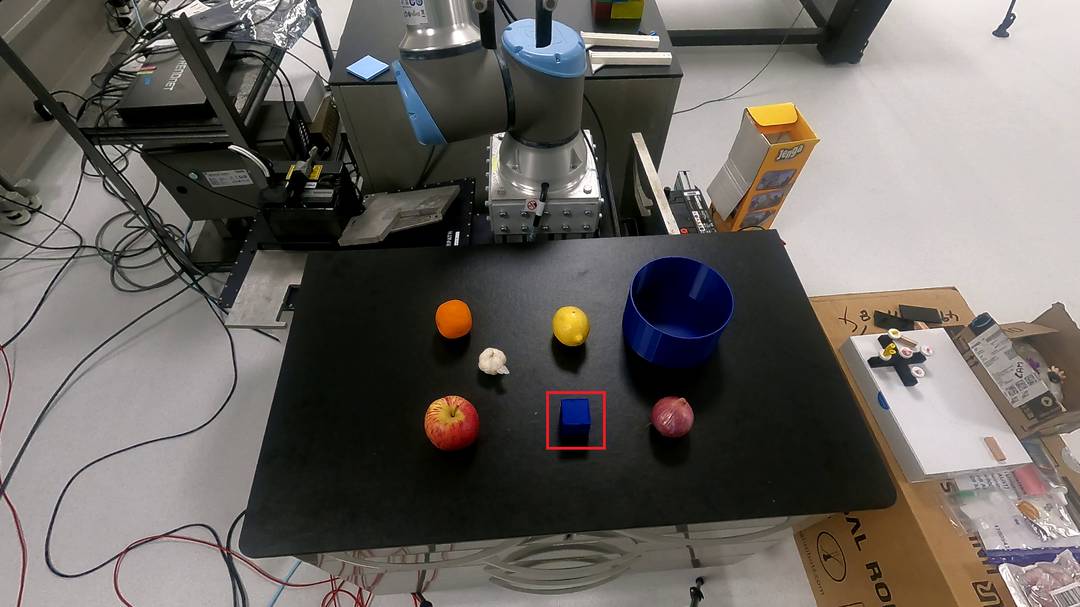}
  \label{fig:experiment_1_7}
  }
  \subfigure[] {
  \includegraphics[trim={0 0 0 0}, clip, width=0.225\textwidth]{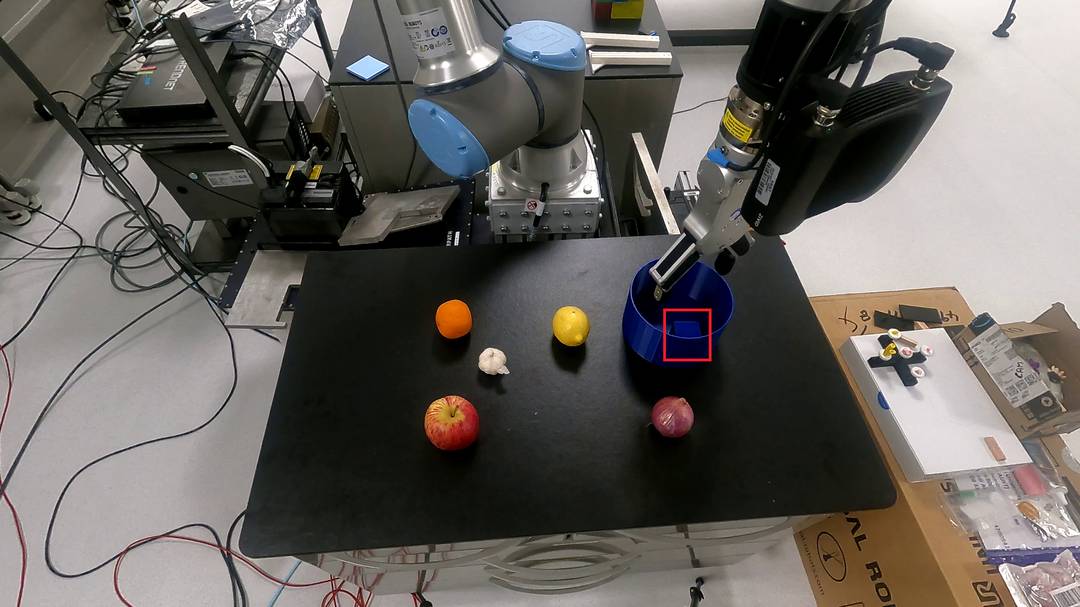}
  \label{fig:experiment_1_8}
  }
  \subfigure[] {
  \includegraphics[trim={0 0 0 0}, clip, width=0.225\textwidth]{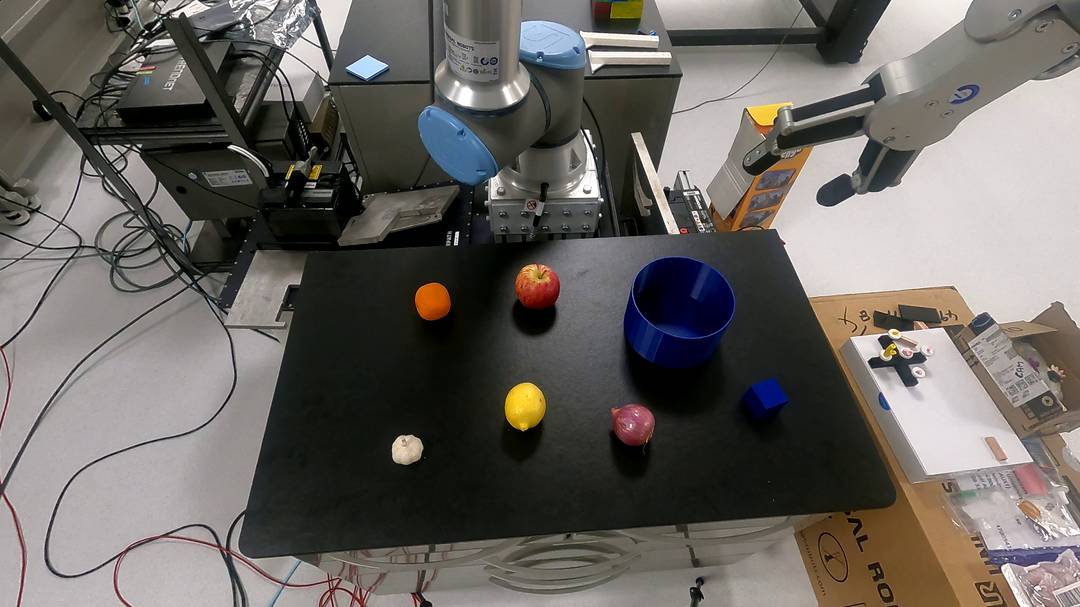}
  \label{fig:experiment_1_9}
  }
  \subfigure[] {
  \includegraphics[trim={0 0 0 0}, clip, width=0.225\textwidth]{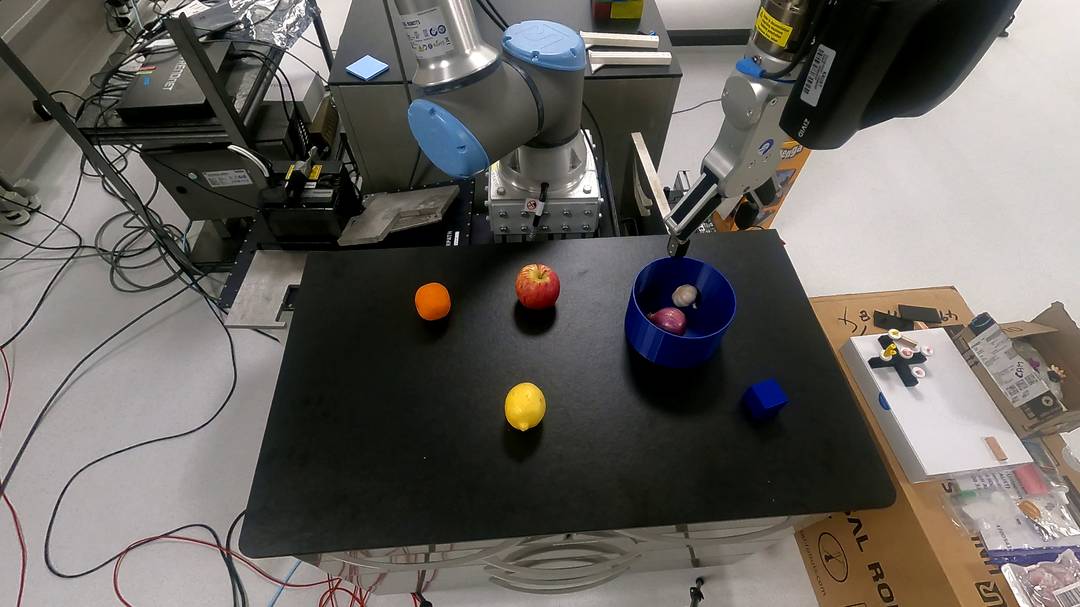}
  \label{fig:experiment_1_10}
  }
  \vspace{-.5em}
  \caption{Experiments I-C and I-D. I-C: (a)-(b) The highlighted non-edible object is selected as the odd one out; I-D: (c)-(d) the robot picks only the ingredients required for fried noodles.}
  \label{fig:experiment_I-C_I-D}
  \vspace{-1.5em}
\end{figure*}

This experiment evaluates whether natural language instructions can be effectively translated into concrete actions requiring recognition, positioning, and contextual reasoning.
The following tasks are designed to evaluate the framework's performance:

\begin{itemize}
\item  \textit{Experiment I-A (Figs.~\ref{fig:experiment_1_1}-\ref{fig:experiment_1_3}):} Relative positioning tasks (e.g., ``\texttt{\small Move the orange between the apple and yarn}'') are designed to assess the stability of the LLM-VLM dialogue and the VLM’s spatial localization capability, particularly its handling of the \textit{in-between} relation. In this setup, the LLM queries the VLM for a point located between the apple and the yarn. The VLM returns the coordinates of the point in the transformed spatial frame, which the LLM uses to update the scene graph and initiate the corresponding pick-and-place operation to move the orange to the designated location.

\item  \textit{Experiments I-B1 and I-B2 (Figs.~\ref{fig:experiment_1_4}-\ref{fig:experiment_1_6}):} Semantic clustering tasks evaluate the framework's capability for understanding the semantics. In Exp. I-B1, the user request ``\texttt{\small Move the lone isolated fruit near the other fruits}'' tests performance under minimal context. In Exp. I-B2, the request ``\texttt{\small The vegetables and fruits are grouped together respectively. But one fruit is isolated. Move it close to where it belongs}'' examines how the performance improves as the user provides additional context. In particular, the model’s capability---or lack thereof---can be observed in how well it understands the notions of ``\texttt{\small lone isolated fruit},'' ``\texttt{\small other fruits},'' and the spatial relation ``\texttt{\small near}'' in Exp.~I-B1, as well as the intended meaning behind ``\texttt{\small move it closer to where it belongs}'' in Exp.~I-B2.

\item \textit{Experiments I-C (Figs.~\ref{fig:experiment_1_7}-\ref{fig:experiment_1_8}) and I-D (Figs.~\ref{fig:experiment_1_9}-\ref{fig:experiment_1_10}):} Context-based manipulation tasks are conducted to evaluate the framework's capabilities in outlier detection (e.g., ``\texttt{\small Transfer the mismatched item from the table to the container}'') and recipe-based selection (e.g., ``\texttt{\small Move the available ingredients for fried noodles into the ingredients box}'').

\end{itemize}
Multiple semantic variations of the user request are employed to evaluate the framework’s contextual understanding, while the initial positions of the manipulated objects are varied in each iteration to create diverse experimental scenarios.

\subsection{Experiment II: Performance Evaluation in Structured Benchmark-Inspired Scenarios}

\begin{figure*}[t]
  \centering
  \subfigure[] {
  \includegraphics[trim={20 0 15 0}, clip, width=0.145\textwidth]{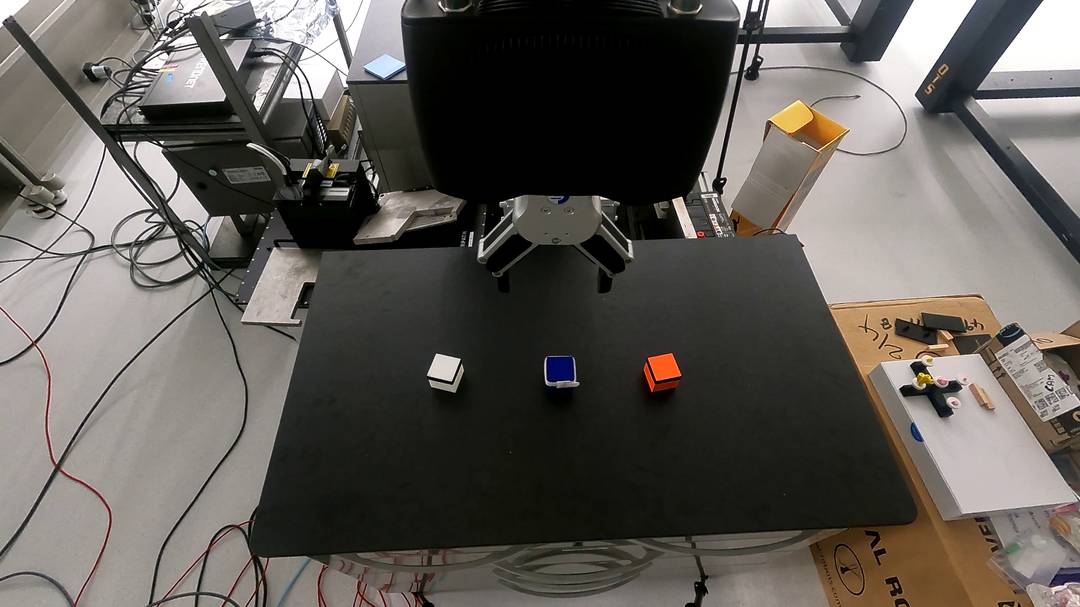}
  \label{fig:experiment_2_1}
  }
  \subfigure[] {
  \includegraphics[trim={20 0 15 0}, clip, width=0.145\textwidth]{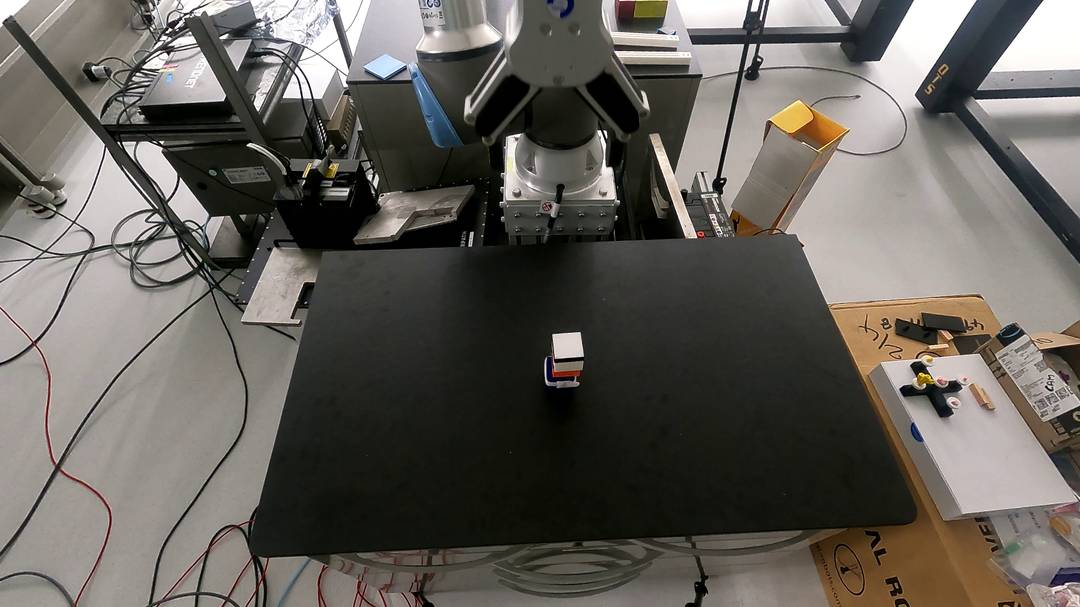}
  \label{fig:experiment_2_2}
  }
  \subfigure[] {
  \includegraphics[trim={16 0 11 0}, clip, width=0.145\textwidth]{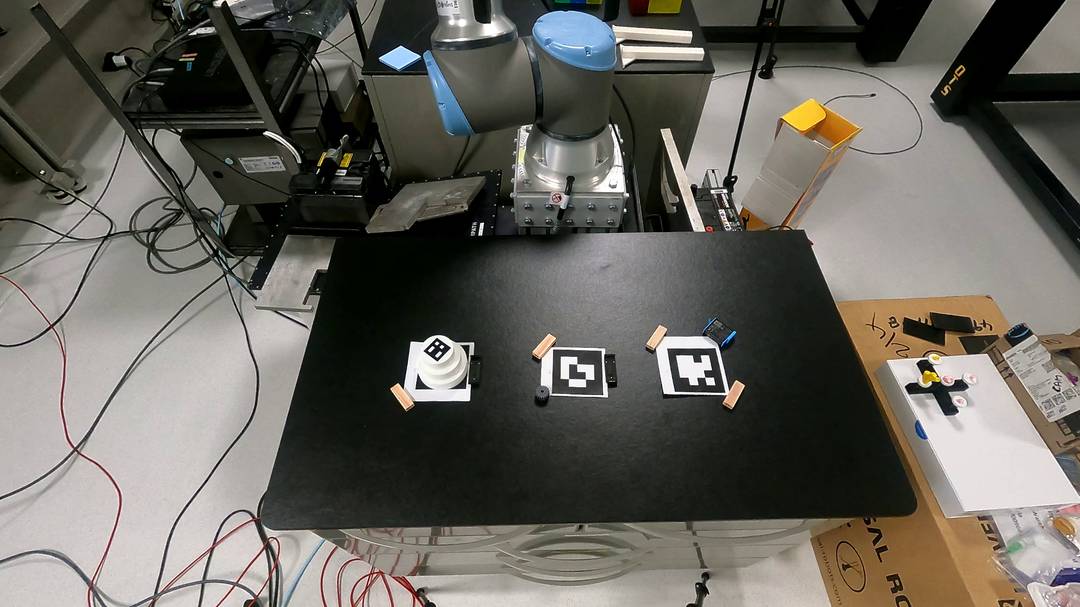}
  \label{fig:experiment_2_3}
  }
  \subfigure[] {
  \includegraphics[trim={16 0 11 0}, clip, width=0.145\textwidth]{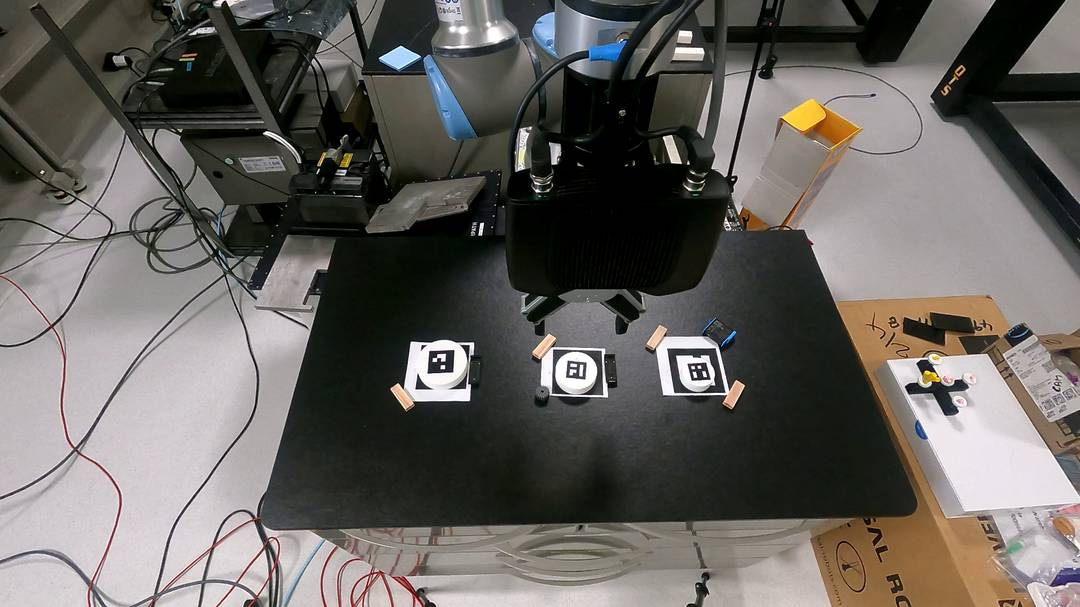}
  \label{fig:experiment_2_5}
  }
  \subfigure[] {
  \includegraphics[trim={16 0 11 0}, clip, width=0.145\textwidth]{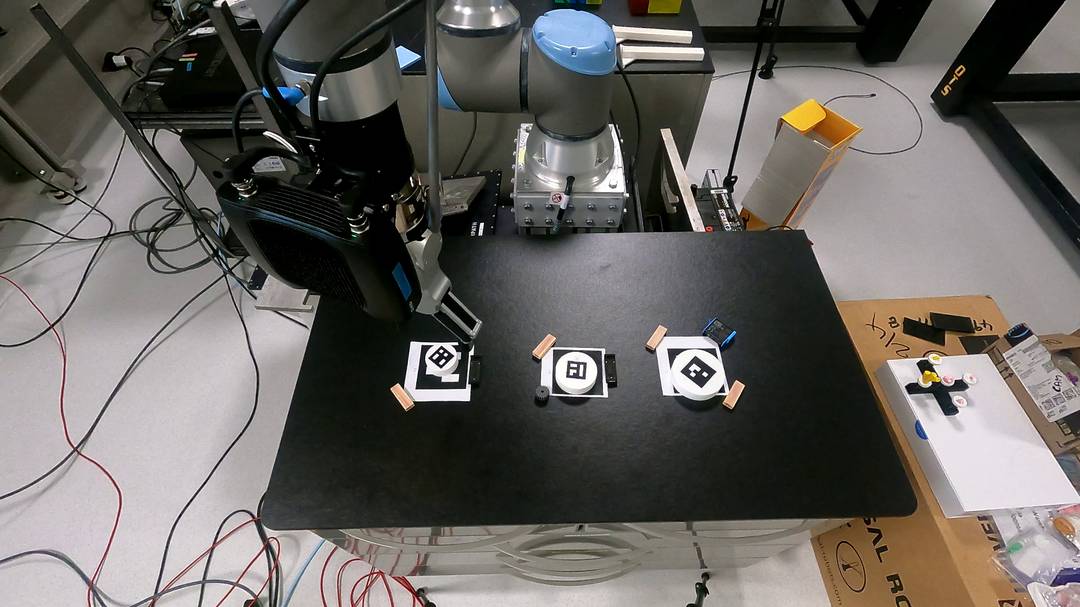}
  \label{fig:experiment_2_8}
  }
  \subfigure[] {
  \includegraphics[trim={16 0 11 0}, clip, width=0.145\textwidth]{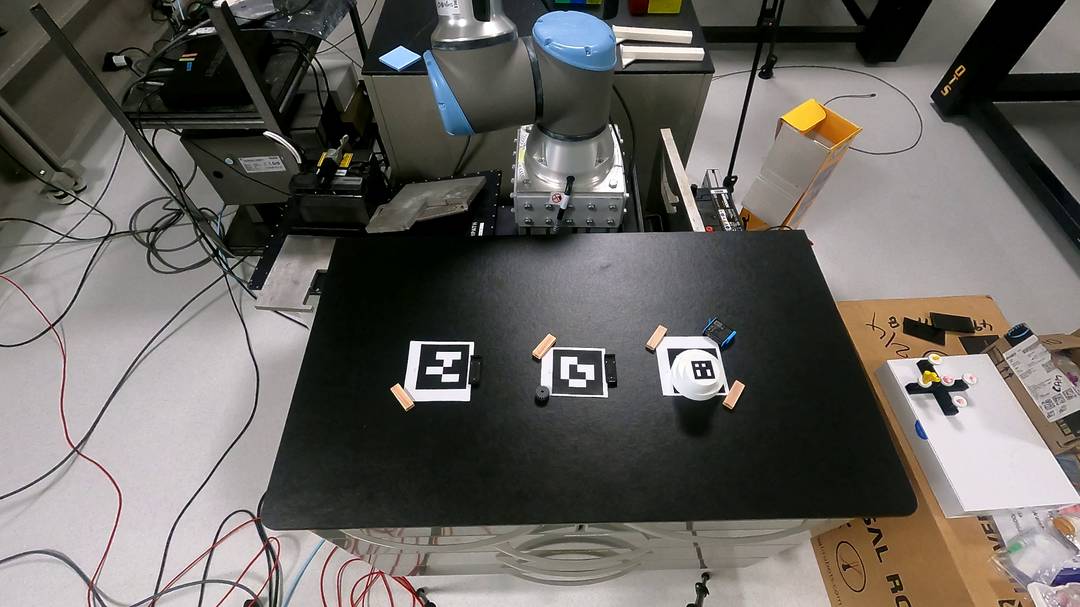}
  \label{fig:experiment_2_10}
  }
  \vspace{-.5em}
  \caption{Experiments II-A and II-B. II-A: (a)–(b) The blocks progress from their initial arrangement on the table to a fully stacked structure. II-B: (c)-(f) The robot solves the Tower of Hanoi puzzle, moving discs step by step from one base to another adhering to the rules of the game.}
  \label{fig:experiment_II-A_II-B}
  \vspace{-1.0em}
\end{figure*}

To evaluate the framework on tasks that require precise robotic reasoning and planning, the following two structured tasks (see Fig.~\ref{fig:experiment_II-A_II-B}) are adopted:

\begin{itemize}

\item \textit{Experiment II-A (Figs.~\ref{fig:experiment_2_1}-\ref{fig:experiment_2_2}):} Block stacking, designed to test iterative spatial reasoning and scene-graph updating. User requests range from straightforward instructions such as ``\texttt{\small Stack the blocks with the one in center as base}'' and ``\texttt{\small I want you to stack the other blocks on top of the white block}'' to more abstract directives like ``\texttt{\small Build something tall using these blocks},'' allowing us to observe the model’s ability to interpret user intent---including abstract or open-ended instructions---and adapt its actions accordingly.

\item \textit{Experiment II-B (Figs.~\ref{fig:experiment_2_3}-\ref{fig:experiment_2_10})}: Tower of Hanoi, a long-horizon puzzle requiring constraint-aware action sequencing. Multiple start-end configurations are tested. To specifically evaluate the cognitive and execution layers, perception was simplified by using AprilTags to identify the base and discs, thereby isolating these layers. Strict adherence to the Tower of Hanoi rules is essential to ensure that each move remains valid and the puzzle’s logical constraints are faithfully respected.

\end{itemize}

Both tasks are successfully completed across multiple trials, demonstrating the framework's ability to generate coherent action sequences through language interpretation and scene-graph–based reasoning. This underscores its capability to operate in structured environments that require step-by-step logical planning and reliable scene-graph update.

\subsection{Experiment III: Advanced Reasoning with Scene Graphs}

\begin{figure}[t]
  \centering
  \subfigure[] {
  \includegraphics[trim={0 0 0 0}, clip, width=0.225\textwidth]{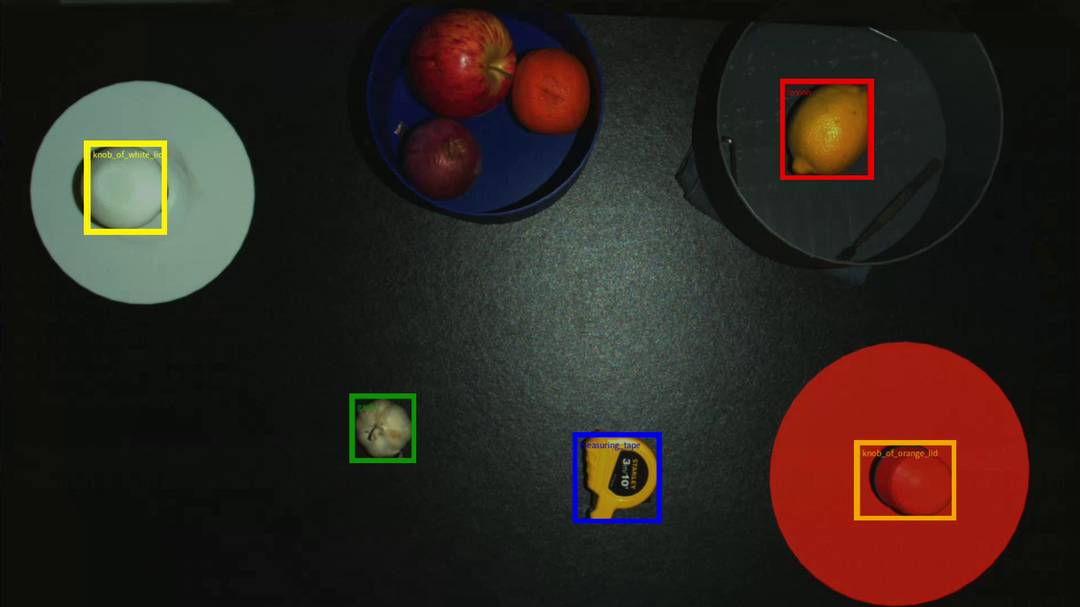}
  \label{fig:experiment_3_9}
  }
  \subfigure[] {
  \includegraphics[trim={0 0 0 0}, clip, width=0.225\textwidth]{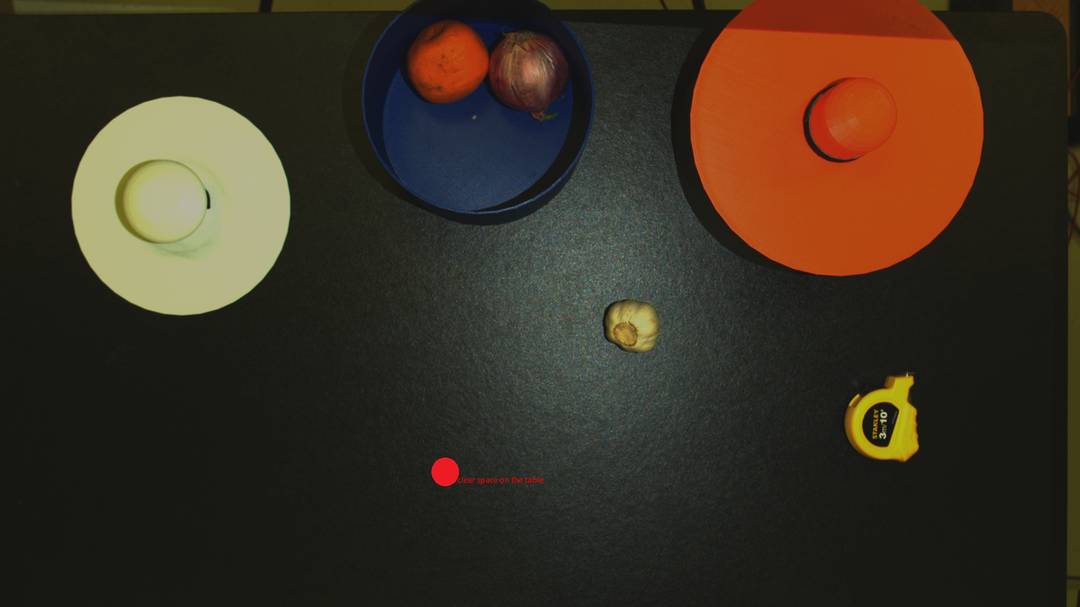}
  \label{fig:experiment_3_10}
  }
  \vspace{-.5em}
  \caption{(a) Yellow bounding boxes show the VLM’s ability to localize fine-grained affordances, such as lid knobs. (b) The bright red point illustrates the VLM’s spatial awareness in assigning a temporary placement location for the lid.}
  \label{fig:experiment_III-A_VLM}
  \vspace{-.5em}
\end{figure}

\begin{figure}[t]
  \centering
  \subfigure[] {
  \includegraphics[trim={5 0 0 0}, clip, width=0.225\textwidth]{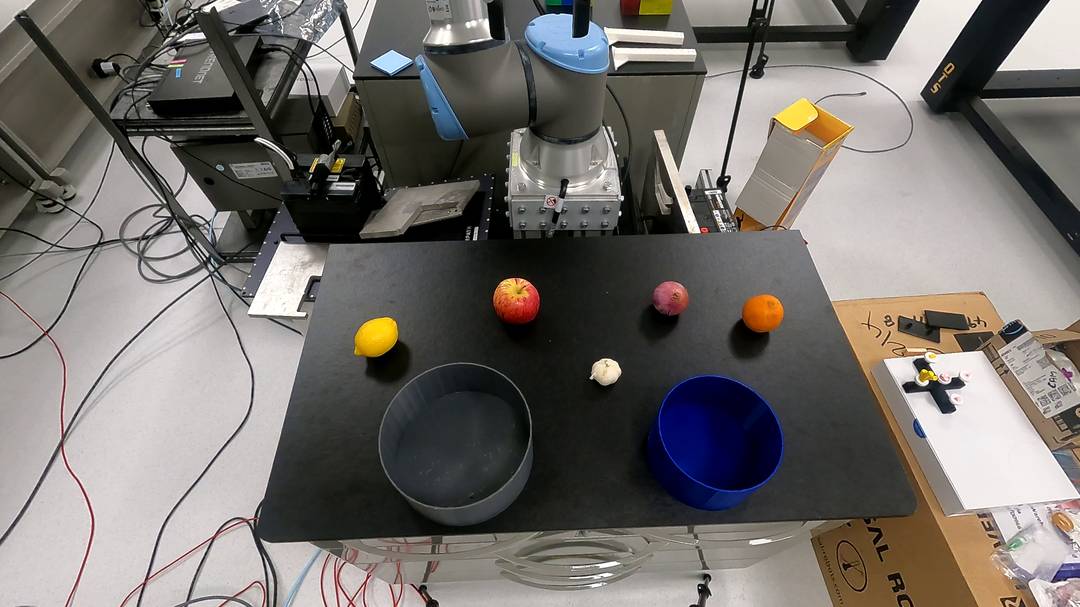}
  \label{fig:experiment_3_1}
  }
  \subfigure[] {
  \includegraphics[trim={5 0 0 0}, clip, width=0.225\textwidth]{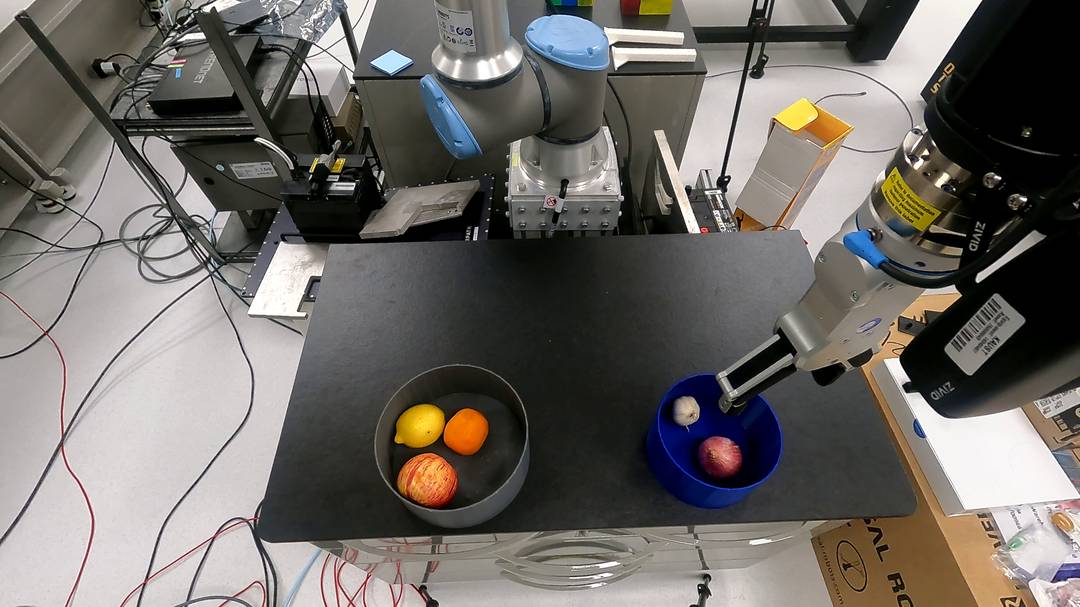}
  \label{fig:experiment_3_2}
  }
  \vspace{-.5em}
  \caption{Experiment III-A. (a)–(b) Items are sorted into two boxes according to inferred categories (fruits vs. vegetables), with fruits placed in the larger container and vegetables in the smaller one.}
  \label{fig:experiment_III-A}
  \vspace{-1.5em}
\end{figure}

\begin{figure*}[t]
  \centering
  \subfigure[] {
  \includegraphics[trim={14 0 11 0}, clip, width=0.145\textwidth]{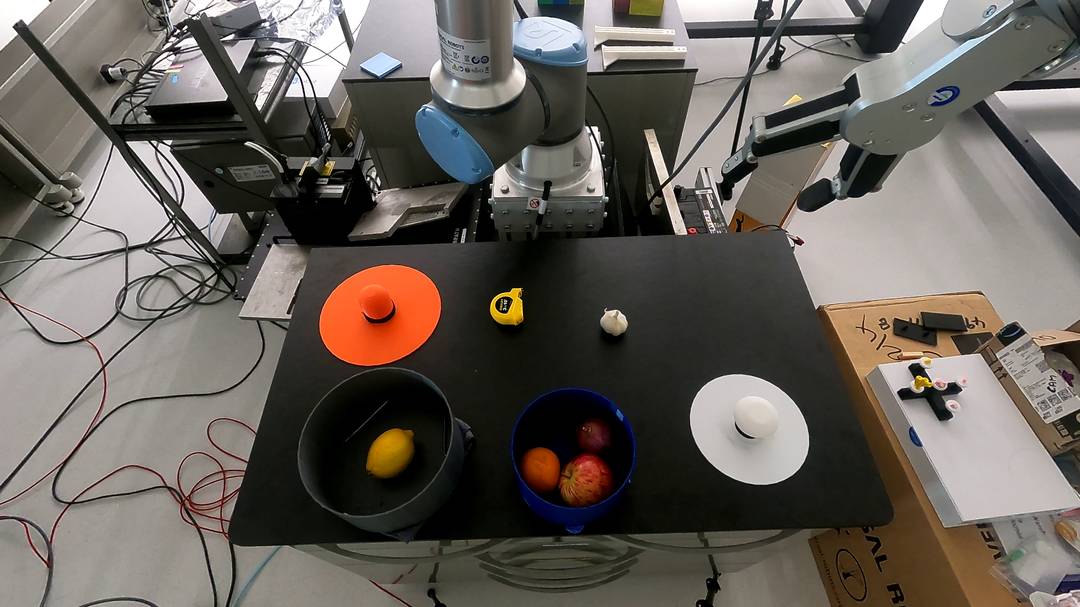}
  \label{fig:experiment_3_3}
  }
  \subfigure[] {
  \includegraphics[trim={14 0 11 0}, clip, width=0.145\textwidth]{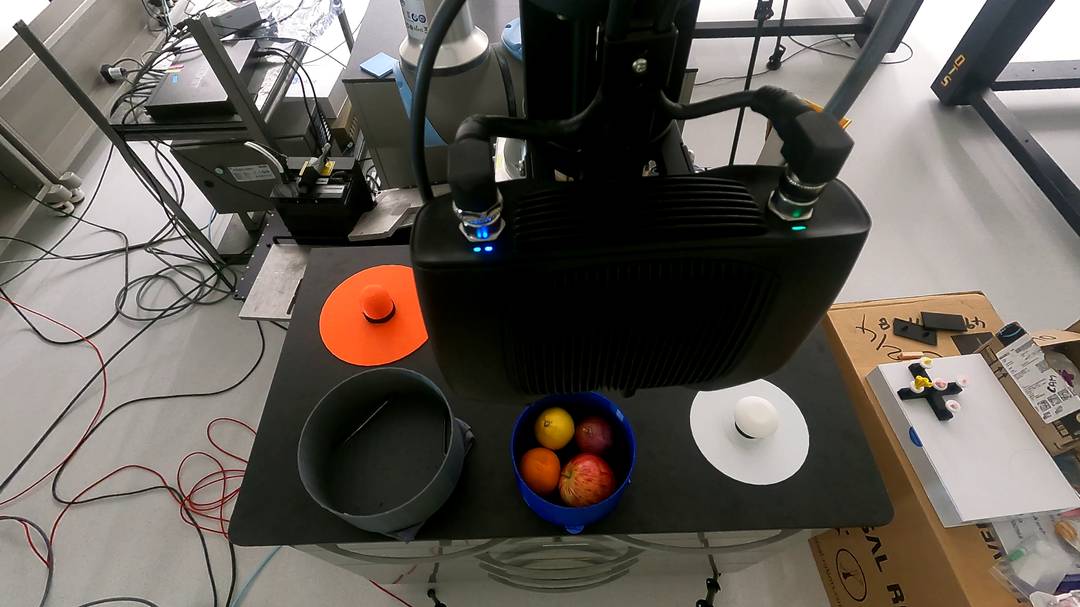}
  \label{fig:experiment_3_4}
  }
  \subfigure[] {
  \includegraphics[trim={14 0 11 0}, clip, width=0.145\textwidth]{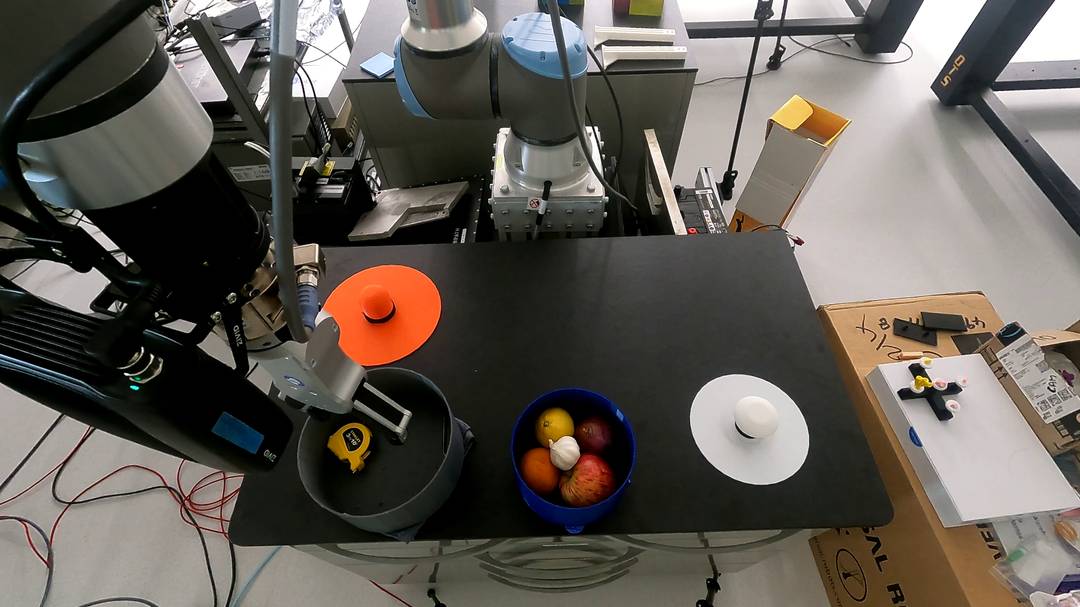}
  \label{fig:experiment_3_5}
  }
  \subfigure[] {
  \includegraphics[trim={14 0 11 0}, clip, width=0.145\textwidth]{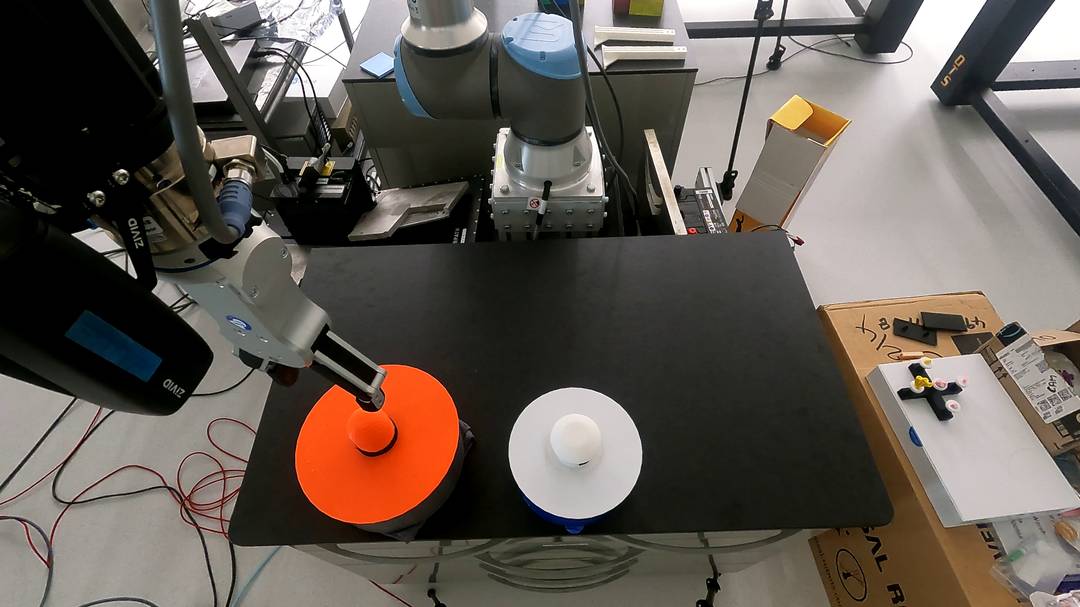}
  \label{fig:experiment_3_6}
  }
  \subfigure[] {
  \includegraphics[trim={14 0 11 0}, clip, width=0.145\textwidth]{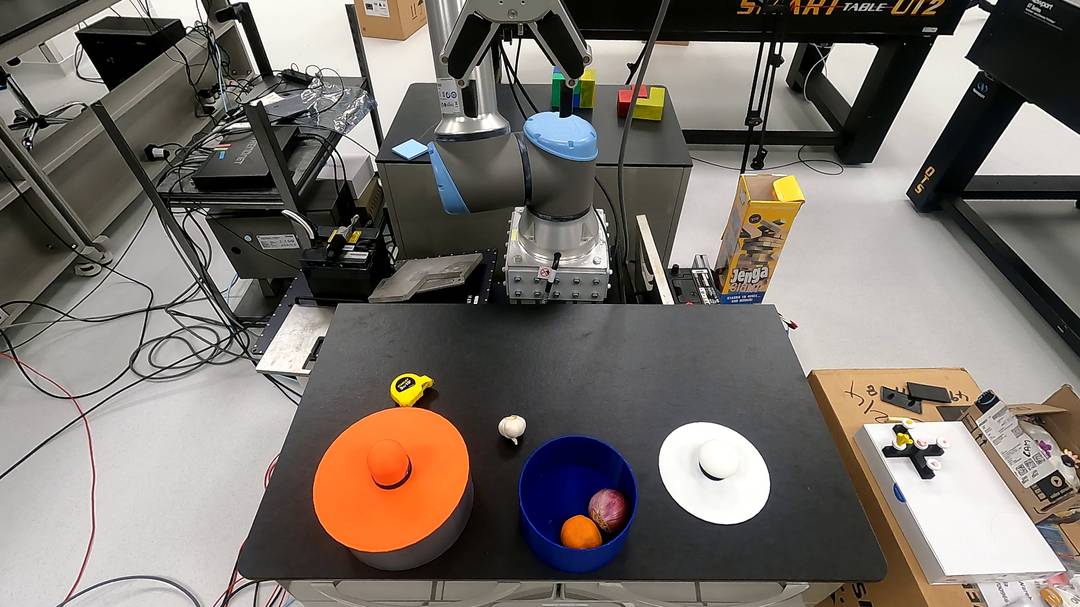}
  \label{fig:experiment_3_7}
  }
  \subfigure[] {
  \includegraphics[trim={14 0 11 0}, clip, width=0.145\textwidth]{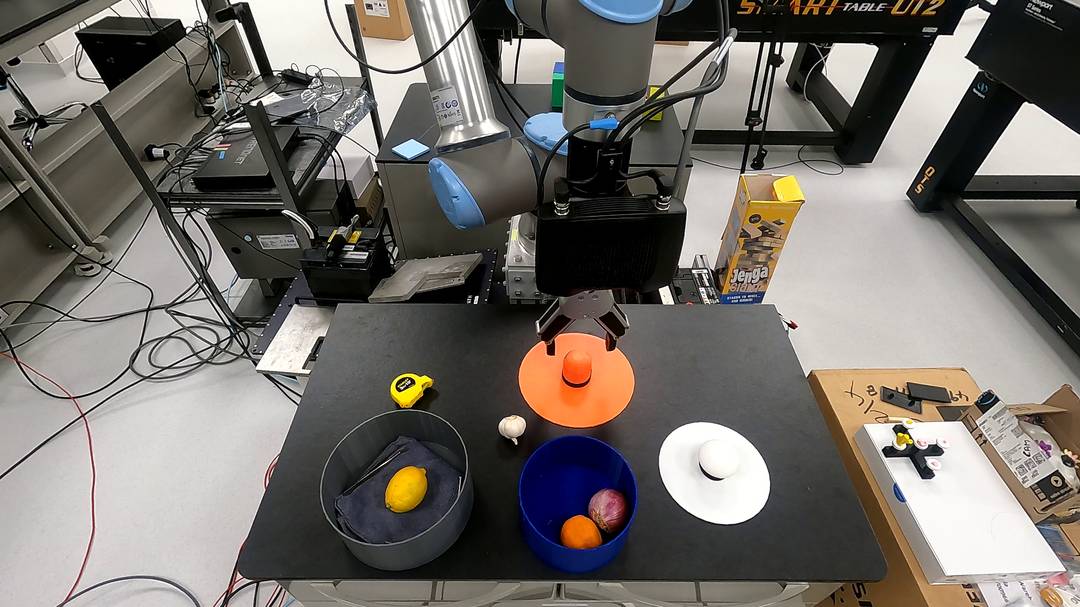}
  \label{fig:experiment_3_8}
  }
  \caption{Experiments III-B and III-C. III-B: (a)–(b) The misplaced lemon is transferred from the toolbox to the food items box; (c) the remaining table objects are organized, (d) and each box is closed with its appropriate lid. III-C: (e)–(f) The robot interprets occlusion by opening the initially closed toolbox, placing the lid at a VLM-identified temporary point, and then proceeding as in III-B.}
  \label{fig:experiment_III-B_III-C}
  \vspace{-1.5em}
\end{figure*}

Further evaluation targets open-ended scenarios requiring multi-step reasoning and semantic categorization (Exps.~III-A$\sim$III-C) as well as occlusion handling (Exp.~III-C). In both Exps.~III-B and III-C, the framework must additionally leverage  the VLM's generalization capability to localize the lid knobs for grasping, as illustrated in Fig.~\ref{fig:experiment_III-A_VLM}.

The tasks include:
\begin{itemize}

\item \textit{Experiment III-A (Figs.~\ref{fig:experiment_3_1}-\ref{fig:experiment_3_2}):} Autonomous sorting, where items are grouped into inferred categories (e.g., fruits vs. vegetables). Two containers are provided: a large box and a small box. 
Because the table contains more fruits than vegetables, the proposed framework is expected to (1) correctly distinguish fruits from vegetables, (2) allocate fruits to the large box, and (3) allocate vegetables to the small box. This setup evaluates both categorization accuracy and container assignment based on relative quantities--all from the simple user input of ``\texttt{\small Put the objects into boxes in an organized manner.}''

\item \textit{Experiment III-B (Figs.~\ref{fig:experiment_3_3}-\ref{fig:experiment_3_6}):} In the autonomous table-organization task, the scene graph includes the grey box and the blue box (positioned on the left and right of the table, respectively), labeled as a toolbox and a food-items box---objects and labels added manually. At the start of the experiment, a lemon is intentionally misplaced in the toolbox (Fig.~\ref{fig:experiment_3_3}). The framework must detect this inconsistency and relocate the lemon to the food-items box (Fig.~\ref{fig:experiment_3_4}) before proceeding to place all remaining objects into their designated boxes (Fig. ~\ref{fig:experiment_3_5}) and closing each with its corresponding lid (Fig. \ref{fig:experiment_3_6}). With only the user command ``\texttt{\small Organize the table}'', the framework is required to infer the full task sequence by extracting relevant details from the initial scene graph, generating and executing a plan, and updating the scene graph online throughout execution.

\item \textit{Experiment III-C (Figs.~\ref{fig:experiment_3_7}-\ref{fig:experiment_3_8}):} This experiment extends Exp.~III-B by introducing an occlusion scenario in the autonomous table-organization task. Initially, the toolbox is closed (Fig.~\ref{fig:experiment_3_7}). The framework must interpret this context via the scene graph, and then leverage the VLM to identify a temporary location on the table (Fig.~\ref{fig:experiment_3_10}) for placing the toolbox lid (Fig.~\ref{fig:experiment_3_8}). The scene graph is then updated to record the lid's temporary position, allowing the framework to effectively handle the occlusion. The subsequent tasks mirror Exp.~III-B: relocating objects to their designated boxes, organizing the table, and closing the boxes with their lids. As before, the only user input is the high-level instruction ``\texttt{\small Organize the table}'', with the system autonomously inferring and executing all steps while continuously updating the scene graph during execution.

\end{itemize}

\subsection{Analysis and Discussions}
\begin{table}[t]
\caption{Experimental Results {(\footnotesize $PF$: Planning Feasibility, \\ $TCR$: Task Completion Rate, $SGH$: Scene Graph Handling.)}\vspace{-1em}}
\begin{center}
\begin{tabular}{|c|c|c|c||c|c|c|c|}
\hline
\textbf{Exp.} & \textbf{$\substack{PF \\(\%)}$} & \textbf{$\substack{TCR \\ (\%)}$} & \textbf{$\substack{SGH \\ (\%)}$} &
\textbf{Exp.} & \textbf{$\substack{PF \\(\%)}$} & \textbf{$\substack{TCR \\ (\%)}$} & \textbf{$\substack{SGH \\ (\%)}$} \\
\hline
\hline
I-A    & 100 & 100 & 100 & II-A   & 100 & 100 & 100 \\
\hline
I-B1 & 100 & 20 & 100 & II-B   & 100 & 100 & 100 \\
\hline
I-B2& 100 & 100 & 100 & III-A  & 100 & 100 & 100\\
\hline
I-C    & 100 & 100 & 100 & III-B  & 95 & 75 & 100 \\
\hline
I-D    & 100 & 80 & 100 & III-C  & 80 & 60 & 100 \\
\hline
\end{tabular}
\label{tab:results}
\end{center}
\vspace{-2em}
\end{table}

We conducted 10 trials for Exp.~I-A, 20 for Exp.~III-B, and 5 trials each for all remaining experiments. As summarized in Table~\ref{tab:results}, the framework consistently achieves high planning feasibility ($PF$) and perfect scene-graph handling ($SGH$). Specifically, $PF \geq 95\%$ in all experiments except Exp. III-C (80\%), highlighting the difficulty of addressing occlusion scenarios, while $SGH$ remains at $100\%$ across all cases. Task completion rate ($TCR$) reflects both the clarity of user instructions and the complexity of scene understanding: fundamental and structured tasks such as Exps.~I-A, I-B2, I-C, II-A, II-B, and III-A reach $TCR = 100\%$, whereas Exp.~I-D shows a moderate decrease to $80\%$. In contrast, the underspecified instructions in Exp.~I-B1 limited the VLM's performance, as the LLM could not provide sufficient context for spatial reasoning, resulting in a $TCR$ of only $20\%$ despite $PF = 100\%$. Rephrasing the user input with richer context in Exp.~I-B2 restored higher $TCR$. 

The most challenging scenarios arise in cluttered or occluded scenes in Exp.~III, where Exp.~III-B achieves $TCR$ of $75\%$ and Exp.~III-C drops further to $60\%$. Long-horizon tasks such as the Tower of Hanoi (Exp.~II-B) and occlusion handling (Exp.~III-C) further highlight the benefit of decoupling extensive planning—managed by a reasoning LLM—from execution—handled by a non-reasoning LLM. Overall, across fundamental capabilities (Exps.~I-A$\sim$I-D), structured benchmarks (Exps.~II-A and II-B), and advanced reasoning (Exps.~III-A$\sim$III-C), the framework reliably maintains robust world-model representations, with failures concentrated in cases of ambiguous language and increased physical complexity. As a result, it achieves high $TCR$ in most experiments. The few VLM errors---limited to Exp. I-B1, where the LLM could not supply adequate contextual cues, and Exp. III-C, where the scene was too visually cluttered to parse---illustrate these two sources of difficulty.

\textbf{Strengths:} Leveraging scene graphs enabled multi-step planning through context retrieval from the graphs, allowing the framework to better interpret high-level user requests (e.g., ``\texttt{\small Organize the table}'' in Exps.~III-B and III-C). The VLM demonstrated strong generalization to novel objects and affordances (e.g., lid-knob localization and temporary placement points) without retraining. Furthermore, decoupling reasoning from execution produced coherent plans while ensuring reliable, deterministic low-level control, with execution calls handled by a non-reasoning model in a feedback loop.

\textbf{Limitations:} The framework remains sensitive to ambiguous language: sparse user requests occasionally caused the LLM to issue queries to the VLM without sufficient context, leading to errors (as observed in Exp.~I-B1). This illustrates error propagation between layers, particularly when performing spatial localization through the VLM. At the execution level, failures also occasionally arose from collisions with manipulatable objects, which were not modeled as dynamic obstacles, thereby reducing $TCR$ in cluttered scenes (Exp.~III-B).

\textbf{Takeaway:}
The layered design of the framework offers a practical middle ground---more adaptable than dataset-heavy VLA systems and more robust than LLM-only and VLM-only planners, particularly in long-horizon, semantically constrained manipulation.

\section{Conclusion} \label{sec:conclusion}

This study introduced a novel foundation-model framework that integrates multiple models across different layers with a scene graph representation of the environment, enabling natural-language manipulation without task-specific training. Experiments demonstrated near-ceiling performance in planning and scene-graph handling, strong perceptual generalization, and reliable execution in both structured and open-ended tasks. Remaining bottlenecks include handling linguistic ambiguity and mitigating collisions arising from the lack of dynamic obstacle modeling at the execution level. As future research, we aim to extend the framework with dexterous object manipulation and to further explore the use of VLMs for manipulating deformable objects and those with complex geometries.

\appendix

Here, we provide only excerpts of the prompts used in our framework, with ``\texttt{\small …}’’ indicating omitted details. The complete prompt specifications are available in Supplementary Materials.

\begin{itemize}
\item {GPT-4.1:} ``\texttt{\small You are a 6-DoF UR10e arm with a two-finger gripper ... Stop if you encounter any failures and let the user know.}''  Another variation of this prompt is used for Exp.~II-B to use AprilTags instead of VLMs.

\item To prevent undesired behaviors, rules are embedded in the system prompt and tool descriptions. For example, the system prompt includes the rule ``\texttt{\small When robot movement is involved (pick or place), you may execute only one movement based tool call at a time.}'' so that GPT can keep track of feedback from the motion planner. Another example from a tool's description is ``\texttt{\small do not call this function after pick and before place}'', which is necessary because an object in the end effector would obscure the camera. When GPT forwards a request to Gemini, these rules are passed along with the available tool definitions.

\item {Gemini 2.5 Pro:} ``\texttt{\small You are a robotic arm. Provide the correct action sequence ... [Available Tools] [Initial Scene Graph].}'' Gemini also gets rules for generating the exhaustive plan, for example: ``\texttt{\small When using scan\_and\_update\_coordinates\_in\_scene\_graph, scan as many VISIBLE objects at once...}''

\item {Qwen2.5 VL:} 
To Retrieve Bounding Box: (1) ``\texttt{\small Do you see [object]? Answer 1 or 0.}'' (2) ``\texttt{\small Output coordinates of all objects in JSON.}''  

To Retrieve Specific Point: ``\texttt{\small Point to [request from GPT] and output a single coordinate in XML.}''
\end{itemize}

\textbf{Model Communication:}
\begin{itemize}
    \item GPT-4.1 $\leftrightarrow$ Gemini 2.5 Pro: GPT forwards the user request—along with the scene graph and available tools—to Gemini and receives an exhaustive plan for task execution.
    \item GPT-4.1 $\leftrightarrow$ Qwen 2.5 VL: GPT prompts Qwen for coordinates or object names to localize and update scene-graph positions; Qwen also answers VQA queries. No chatlog is stored on Qwen.
    \item GPT-4.1 $\leftrightarrow$ Motion Primitives / Scene Graph: GPT passes parameters to motion or scene-graph functions and receives success/failure feedback.
    \item A parameter is an input a tool uses to operate—like object\_name in pick(object\_name). The function returns text for VQA with VLM, or a success/failure message (with a reason if it fails) for tasks like object localization, scene graph editing, or motion planning.

\end{itemize}

\textbf{Available Tools for GPT 4.1:} 
\begin{itemize}
    \item {Motion Primitives:} pick object and place object, implemented with Nvidia cuRobo
    \item {Perception:} ask VQA VLM, scan and update coordinates in scene graph (Given a set of object names, detect visible objects, localize their 3D poses using VLM issued bounding boxes, and update the scene graph accordingly.), get a specific coordinate point using VLM (GPT provides a prompt and VLM returns the coordinate of the points satisfying that prompt. e.g. point in-between object 1 and object 2). Points are typically used in our implementation to get locations to place objects. However, in Exp.~II-A, points were used for both picking and placing since the scan and update tool was not provided to LLMs. For Exp.~II-B we use get current position of visible AprilTags instead of VLM based functions.
    \item {Scene Graph:} The scene graph can be modified using two key tools: \texttt{add object to scene graph (object name, attributes)} to insert a new object with the specified attributes, and \texttt{edit scene graph(object name, attribute name, value)} to update an existing object's attributes, such as containment relationships. For example, if the robot removes an object from a box and the box becomes empty, GPT would update the scene graph with \texttt{edit\_scene\_graph("box", "contains", None)}.
    \item {Cognitive Layer:} Task sequence planning using a reasoning model (Gemini 2.5 Pro).
\end{itemize}

\textbf{Failure Handling:}  
GPT tracks every function call and feedback. If a step fails, it alerts the user, re-plans, and retries. After repeated failures, it may suggest skipping the object or ask the user to reposition it when it’s not visible—something we experienced and followed in Exps.~III-B and III-C.


\bibliographystyle{IEEEtran}
\bibliography{references}

\twocolumn[
  \begin{@twocolumnfalse}
  \begin{center}
  {\LARGE\bfseries Supplementary Materials}
  \end{center}
  \vspace{1em}
  \end{@twocolumnfalse}
]

\setcounter{section}{0}
\def\thesection{\Roman{section}}
\def\thesectiondis{\Roman{section}.}  
\def\thesubsection{\Alph{subsection}}

\makeatletter
\def\section{\@startsection{section}{1}{\z@}{1.5ex plus 1.5ex minus 0.5ex}%
{0.7ex plus 1ex minus 0ex}{\normalfont\normalsize\centering\scshape}}
\setcounter{secnumdepth}{3}
\makeatother

\section{Prompt used for GPT 4.1}

\subsection{GPT 4.1}

The following system prompt is used across all experiments for GPT-4.1.

``You are a 6DoF casual and friendly Universal Robotics UR10e arm with a two finger gripper. Your task is to initiate the conversation and get a request from the user first. Then, pass it to the \texttt{plan\_using\_advanced\_llm} which can give you the right order of tool calls with parameters. Don't ask questions to the user and simply pass his request to the advanced LLMs since they also have access to the scene graph and available tools. Based on the output of \texttt{plan\_using\_advanced\_llm} planner you MUST STRICTLY ADHERE to its plan and execute its plan. When robot movement is involved (pick or place), you may execute only one movement based tool call at a time. Pick, modifying the scene graph can be called at once but you cannot call pick and place at once. Observe the feedback before going to the next manipulation/movement tool call. Once you receive the plan, ask user for confirmation. Once he confirms you can execute until everything is done. Stop if you encounter any failures and let user know.''

\subsection{Tool Descriptions}

In addition to the aforementioned system prompt for GPT 4.1, the tool descriptions are also added to the system prompt to enable tool use (function calling) with GPT 4.1.

For object manipulation, the following tools were used.

\begin{itemize}
    \item \textbf{\texttt{pick\_object}:} ``Makes the robot pick a provided object. The name MUST precisely match what is in the database/scene graph whose coordinate is available.''
    
    \textbf{Parameter:} \texttt{object\_name}
    
    \item \textbf{\texttt{place\_object}:} ``Makes the robot to place an object in hand safely at the provided place name. The name MUST precisely match what is in the database/scene graph whose coordinate is available.''
    
    \textbf{Parameter:} \texttt{place\_position\_name}
\end{itemize}

For VLM interaction, the following tools were used.

\begin{itemize}
    \item \textbf{\texttt{ask\_vqa\_vlm}:} ``This is a VLM (QwenVL 2.5). You can ask VQA. The VLM will answer with whatever you want to know. Only for single Q\&A and not conversations since the chat history is not stored.''
    
    \textbf{Parameter:} \texttt{query\_to\_vlm}
    
    \item \textbf{\texttt{scan\_and\_update\_coordinates\_in\_scene\_graph}:} ``DO NOT CALL AFTER PICK\_OBJECT. The camera mounted on the robotic arm will scan and update the position of the requested object using a VLM. The camera mounted near gripper needs unobstructed view when scanning and it cannot see hidden objects. Therefore, do not call this function after pick and before place. This function fills/updates the scene graph nodes (targets) with coordinates by itself. This can only scan identifiable individual objects. To simply get points you need to user \texttt{get\_a\_specific\_coordinate\_point\_using\_vlm}.''
    
    \textbf{Parameter:} \texttt{targets\_to\_scan}

    \item \textbf{\texttt{get\_a\_specific\_coordinate\_point\_using\_vlm}:} ``DO NOT CALL AFTER PICK\_OBJECT. The camera mounted on the robotic arm will look at the workspace. Then VLM will give you the specific point you ask for in the workspace. Output given to you will be of the form [x,y,z]. Based on response you need to add/update the scene graph with received output coordinates by yourself. The gripper should also be free of any objects when scanning. Therefore DO NOT call this function after pick\_object and before \texttt{place\_object}. Strictly do NOT add `I want [x,y,z] coordinate' in prompt and keep the prompt SHORT. The camera ALWAYS takes the TOP VIEW Photo of the workspace/table. ''
    
    \textbf{Parameter:} prompt\_to\_vlm
\end{itemize}

For the Tower of Hanoi experiment (Exp. II-B), VLM was not used and the following tool was used instead to localize Apriltags.

\begin{itemize}
    \item \textbf{\texttt{get\_current\_position\_of\_visible\_apriltags}:} ``This function lets you know of the TAG ID and Position [x,y,z] of all the currently visible Apriltags. But ONLY VISIBLE Objects with Apriltags are captured. Obscured objects are not visible. Note that camera captures top down view of table.''
    
    \textbf{Parameter:} \texttt{trigger}
\end{itemize}

For modifying the scene graph, the following tools were used.

\begin{itemize}
    \item \textbf{\texttt{add\_object\_to\_scenegraph}:} ``Adds a new object in the scene graph with specified parameters.''
    
    \textbf{Parameters:} \texttt{object\_name, affordance, position\_in\_cartesian\_space, things\_to\_know, coordinates, contains}
    
    \item \textbf{\texttt{edit\_scenegraph}:} ``Edit the attribute of any node that is already present in the scene graph.''
    
    \textbf{Parameters:} \texttt{node\_name, attribute\_name, value}
\end{itemize}

The following tool was used to send the required data to Gemini.
\begin{itemize}
    \item \textbf{\texttt{plan\_using\_advanced\_llm}:} ``You will get a detailed plan from advanced LLM to execute. This ensures high success rate.''
    
    \textbf{Parameter:} \texttt{request\_from\_user}
\end{itemize}

\section{Prompts used for Gemini 2.5 Pro}

\subsection{Default Prompt}

This is the default prompt used in all the experiments except Exp. II-B.

``You are a robotic arm. Your task is to give the right sequence to achieve the user request  \texttt{[request\_from\_user]}

Important note:
1) If you move (pick or place) an object, you need to update its position (coordinates) again before attempting another pick or place. If you don't do that, the robot will unintentionally approach the past position available in the scene graph.
2) Scanning is time-consuming. So update the position of manipulated objects if and only if you want to manipulate it again which requires the latest position.
3) Update the scene graph as required. But don't waste time in scanning newly updated positions unless you plan to manipulate those objects further.
4) MANDATORY: You are PROHIBITED to use \texttt{get\_a\_specific\_coordinate\_point\_using\_vlm} AFTER \texttt{pick\_object} since an object held in hand will block cameras completely. STRICT RULE. MUST FOLLOW!!.
5) Make sure to mark any placeholder values in case it depends on a previous function call in order for the actual action executing LLM to understand properly.
6) You are far more intelligent (way larger model) than the ones used by VLM. So only use VLM as your eyes and not for anything that involves logic, reasoning and wider knowledge base. Use the VQA and Monologue to perceive---that's it.
7) When using \texttt{scan\_and\_update\_coordinates\_in\_scene\_graph}, scan as many VISIBLE objects at once since scanning one by one can take some time since the robot needs to reach several vantage points to construct pointcloud for processing.

You can use the following functions:
\texttt{[Available Tools]}

The following is the scene graph representation available currently.
\texttt{[Initial Scene Graph]}''

The \texttt{Available Tools} are identical to the tool descriptions provided to GPT 4.1 in Section I-B of Supplimentary Materials. The \texttt{Initial Scene Graph} is the available scene graph in JSON sent to the LLM in text format.

\subsection{Alternative Prompt}
This alternative prompt is used only for the Tower of Hanoi experiment (Exp. II-B) involving Apriltags.

``You are a robotic arm. Your task is to give the right sequence to achieve the user request  \texttt{[request\_from\_user]}

Important note:
1) If you move (pick or place) an object, you need to update its position (coordinates) again before attempting another pick or place. If you don't do that, the robot will unintentionally approach the past position available in the scene graph.
2) NEVER EVER use \texttt{get\_current\_position\_of\_visible\_apriltags} between pick-and-place. Because an object in the end effector will block the view of the workspace. So you can use this function only after placing whatever is in hand.
3) Use the \texttt{get\_current\_position\_of\_visible\_apriltags} function to get the latest position. Make sure to update in the scene graph after fetching the values. So that pick-and-place can use that.
4) Make sure to mark any placeholder values in case it depends on a previous function call in order for the actual action executing LLM to understand properly.
5) Apriltags are used for localization. But Apriltags are only seen by the camera if it is not obscured when capturing the top-down view of the table/workspace. So don't attempt to see potentially obscured objects.
6) When using \texttt{place\_object}, use the name of the object that will be underneath the current object, rather than using generic labels like \texttt{base\_1, base\_2, base\_3} unless you are specifically placing on the base surface.

You can use the following functions:
\texttt{[Available Tools]}

The following is the scene graph representation available currently.
\texttt{[Initial Scene Graph]}''

\section{Prompts used for Qwen2.5 VL}

\subsection{Getting Bounding Box}

The first prompt verifies the presence of a given object to prevent the VLM from hallucinating non-existent ones. This step ensures that the VLM explicitly confirms whether the object is present. Only the objects confirmed to exist are subsequently sent to the VLM for bounding box generation.

\begin{itemize}
    \item Prompt: ``Do you see \texttt{[object name]} in the image. Answer strictly in binary. 1 or 0.''
    \item System Prompt: ``Output is STRICTLY Binary. 1 or 0.''
\end{itemize}

Once the objects present in the frame are confirmed, their bounding boxes can be obtained using the following prompt.

\begin{itemize}
    \item Prompt: ``Outline the position of  \texttt{object names} and output all the coordinates in the JSON format.''
    \item System Prompt: Strictly maintain format
    \begin{lstlisting}[language=json]
    [{"bbox_2d": [integer, integer, integer, integer], "label": "obj_name"},
    {"bbox_2d": [integer, integer, integer, integer], "label": "obj_name"},
    ...
    ]
    \end{lstlisting}

In this case, the system prompt specifies the desired output JSON format required for parsing.
    
\end{itemize}
\subsection{Getting Specific Point}

The system prompt defines the required output format for parsing. The prompt used is shown below.

\begin{itemize}
    \item System Prompt: ``STRICTLY ADHERE TO OUTPUT FORMAT \texttt{<points x y>object</points>}. Single Coordinate. STRICTLY ADHERE TO THIS FORMAT!!!!''
    \item Prompt: ``Point to the \texttt{prompt from GPT}. STRICTLY output a SINGULAR coordinate in XML format \texttt{<points x y>object</points>}''
\end{itemize}

\section{Model Variants and Access}

\begin{itemize}
    \item GPT 4.1 - \texttt{gpt-4.1-2025-04-14}. Accessed via the OpenAI API.
    \item Gemini 2.5 Pro - \texttt{gemini-2.5-pro-preview-05-06}. Accessed via the Google Gemini API.
    \item Qwen2.5-VL - \texttt{32B and 72B}. Accessed via OpenRouter, with the specific variant selected based on availability.
\end{itemize}

\section{Scene Graph Example}

The following is the initial scene graph provided to the LLMs for Exp. III-A.

\begin{lstlisting}[language=json]
{
    "workspace": {
      "affordance": [
        "None"
      ],
      "contains": [
        "table"
      ],
      "position_in_cartesian_space": "irrelevant",
      "things_to_know": "None",
      "coordinates": []
    },
    "table": {
      "affordance": [
        "fixed in space"
      ],
      "contains": [
        "small_box",
        "large_box",
        "orange",
        "apple",
        "lemon",
        "garlic",
        "red_onion"
      ],
      "position_in_cartesian_space": "irrelevant. coordinates not available as table refers to the whole accessible workspace. You need specific point in the table if you want to place something on the table.",
      "things_to_know": "None",
      "coordinates": []
    },
    "orange": {
        "affordance": [
          "pickable",
          "edible"
        ],
        "contains": [],
        "position_in_cartesian_space": "centroid_can_be_obtained",
        "things_to_know": "A small, round, orange-colored fruit.",
        "coordinates": []
      },
      "apple": {
        "affordance": [
          "pickable",
          "edible"
        ],
        "contains": [],
        "position_in_cartesian_space": "centroid_can_be_obtained",
        "things_to_know": "A medium-sized, round fruit with red and yellow striped skin.",
        "coordinates": []
      },
      "lemon": {
        "affordance": [
          "pickable",
          "edible"
        ],
        "contains": [],
        "position_in_cartesian_space": "centroid_can_be_obtained",
        "things_to_know": "A small, oval, yellow-colored fruit.",
        "coordinates": []
      },
      "garlic": {
        "affordance": [
          "pickable",
          "edible"
        ],
        "contains": [],
        "position_in_cartesian_space": "centroid_can_be_obtained",
        "things_to_know": "A small, bulbous, off-white vegetable with a papery outer skin.",
        "coordinates": []
      },
      "red_onion": {
        "affordance": [
          "pickable",
          "edible"
        ],
        "contains": [],
        "position_in_cartesian_space": "centroid_can_be_obtained",
        "things_to_know": "A bulb-shaped vegetable with a deep purple outer layer.",
        "coordinates": []
      },
    "small_box": {
        "affordance": [
          "pickable"
        ],
        "contains": [],
        "position_in_cartesian_space": "Position is explicitly defined",
        "things_to_know": "This is fixed in table. This is a cylindrical box. It has a smaller radius.",
        "coordinates": [0.19957663118839264, -0.6754058599472046, 0.14970232427120209]
      },

      "large_box": {
        "affordance": [
          "pickable"
        ],
        "contains": [],
        "position_in_cartesian_space": "Position is explicitly defined. This is a cylindrical box. It has a larger radius.",
        "things_to_know": "This is fixed in table",
        "coordinates": [-0.17225371301174164, -0.6708526611328125, 0.14970232427120209]
      }
}

\end{lstlisting}

By the end of Exp. III-A, the LLM had made several modifications to the scene graph, as shown below.

\begin{lstlisting}[language=json]
ras
{
  "workspace": {
    "affordance": [
      "None"
    ],
    "contains": [
      "table"
    ],
    "position_in_cartesian_space": "irrelevant",
    "things_to_know": "None",
    "coordinates": []
  },
  "table": {
    "affordance": [
      "fixed in space"
    ],
    "contains": [
      "small_box",
      "large_box"
    ],
    "position_in_cartesian_space": "irrelevant. coordinates not available as table refers to the whole accessible workspace. You need specific point in the table if you want to place something on the table.",
    "things_to_know": "None",
    "coordinates": []
  },
  "orange": {
    "affordance": [
      "pickable",
      "edible"
    ],
    "contains": [],
    "position_in_cartesian_space": "inside large_box",
    "things_to_know": "A small, round, orange-colored fruit.",
    "coordinates": []
  },
  "apple": {
    "affordance": [
      "pickable",
      "edible"
    ],
    "contains": [],
    "position_in_cartesian_space": "inside large_box",
    "things_to_know": "A medium-sized, round fruit with red and yellow striped skin.",
    "coordinates": []
  },
  "lemon": {
    "affordance": [
      "pickable",
      "edible"
    ],
    "contains": [],
    "position_in_cartesian_space": "inside large_box",
    "things_to_know": "A small, oval, yellow-colored fruit.",
    "coordinates": []
  },
  "garlic": {
    "affordance": [
      "pickable",
      "edible"
    ],
    "contains": [],
    "position_in_cartesian_space": "inside small_box",
    "things_to_know": "A small, bulbous, off-white vegetable with a papery outer skin.",
    "coordinates": []
  },
  "red_onion": {
    "affordance": [
      "pickable",
      "edible"
    ],
    "contains": [],
    "position_in_cartesian_space": "inside small_box",
    "things_to_know": "A bulb-shaped vegetable with a deep purple outer layer.",
    "coordinates": []
  },
  "small_box": {
    "affordance": [
      "pickable"
    ],
    "contains": [
      "garlic",
      "red_onion"
    ],
    "position_in_cartesian_space": "Position is explicitly defined",
    "things_to_know": "This is fixed in table. This is a cylindrical box. It has a smaller radius.",
    "coordinates": [
      0.19957663118839264,
      -0.6754058599472046,
      0.1497023242712021
    ]
  },
  "large_box": {
    "affordance": [
      "pickable"
    ],
    "contains": [
      "orange",
      "apple",
      "lemon"
    ],
    "position_in_cartesian_space": "Position is explicitly defined. This is a cylindrical box. It has a larger radius.",
    "things_to_know": "This is fixed in table",
    "coordinates": [
      -0.17225371301174164,
      -0.6708526611328125,
      0.1497023242712021
    ]
  }
}

\end{lstlisting}

The scene graph handling in this case is considered successful, as the hierarchies were updated: the fruits and vegetables that were originally on the table were moved into two boxes. Specifically, \texttt{garlic} and \texttt{red\_onion} were placed in the \texttt{small\_box}, while \texttt{orange}, \texttt{apple}, and \texttt{lemon} were placed in the \texttt{large\_box}.

\end{document}